\documentclass{article}

\usepackage[final]{neurips_2024}

\usepackage[utf8]{inputenc} 
\usepackage[T1]{fontenc}    
\usepackage{hyperref}       
\usepackage{url}            
\usepackage{booktabs}       
\usepackage{amsfonts}       
\usepackage{nicefrac}       
\usepackage{microtype}      
\usepackage{xcolor}         

\usepackage{microtype}
\usepackage{graphicx}
\usepackage{subfigure}
\usepackage{booktabs} 

\usepackage{hyperref}

\usepackage{amsmath}
\usepackage{amssymb}
\usepackage{mathtools}
\usepackage{amsthm}
\usepackage{url}
\usepackage{times}
\usepackage{epsfig}
\usepackage{graphicx}
\usepackage{nccmath}
\usepackage{comment}
\usepackage{microtype}
\usepackage{subfigure}
\usepackage{booktabs} 
\usepackage{wrapfig}
\usepackage{stackengine}
\usepackage{enumitem}
\usepackage{eqparbox,array}
\usepackage{multirow}
\usepackage{makecell}

\usepackage[capitalize,noabbrev]{cleveref}
\def\eg{\textit{e.g}., } 
\def\ie{\textit{i.e}., }

\DeclareMathOperator*{\argmax}{arg\,max}

\def\yenrule{\rule{1.3ex}{.1ex}}
\def\textyen{\renewcommand\stacktype{L}\stackon[.4ex]{\stackon[.65ex]{Y}{\yenrule}}{\yenrule}}

\theoremstyle{plain}

\theoremstyle{definition}

\theoremstyle{remark}

\title{Sample Selection via Contrastive Fragmentation for Noisy Label Regression}

\author{%
  Chris Dongjoo Kim$^{1,2}$\thanks{These authors contributed equally to this work.},\quad Sangwoo Moon$^{1}$\footnotemark[1],\quad Jihwan Moon$^{1}$ \\
  \textbf{Dongyeon Woo}$^{1}$,\quad \textbf{Gunhee Kim}$^{1,2}$ \\
  $^{1}$Seoul National University, $^{2}$LG AI Research \\
  \texttt{\{cdjkim, sangwoo.moon, jihwan.moon, dongyeon.woo\}@vision.snu.ac.kr}\\ 
  \texttt{gunhee@snu.ac.kr} \\
}

\begin{document}

\maketitle

\begin{abstract}
As with many other problems, real-world regression is plagued by the presence of noisy labels, an inevitable issue that demands our attention. 
Fortunately, much real-world data often exhibits an intrinsic property of continuously ordered correlations between labels and features, where data points with similar labels are also represented with closely related features.
In response, we propose a novel approach named ConFrag, where we collectively model the regression data by transforming them into disjoint yet contrasting fragmentation pairs. 
This enables the training of more distinctive representations, enhancing the ability to select clean samples.
Our ConFrag framework leverages a mixture of neighboring fragments to discern noisy labels through neighborhood agreement among expert feature extractors.
We extensively perform experiments on six newly curated benchmark datasets of diverse domains, including age prediction, price prediction, and music production year estimation.
We also introduce a metric called Error Residual Ratio (ERR) to better account for varying degrees of label noise.
Our approach consistently outperforms fourteen state-of-the-art baselines, being robust against symmetric and random Gaussian label noise.\footnote{The code is available at \url{https://github.com/cdjkim/ConFrag}}.

\end{abstract}

\section{Introduction}\label{sec:introduction}
Regression 
is an important task in many disciplines such as finance~\citep{zhang17finance,wu20finance}, medicine~\citep{coen21medicine,tanaka22medicine}, 
economics~\citep{zhang22economics}, physics~\citep{sial2020light, doi22physics}, geography~\citep{liu23elnino} and more.
However, real-world regression labels are prone to being corrupted with noise, making it an inevitable problem to overcome in practical applications. 
In previous research, noisy label regression has been studied much in age estimation with noise incurred from Web data crawling~\citep{rothe18imdb, lin2021imdbclean}.
Beyond that, the issues of continuous label errors have also been reported in the tasks of object detection~\citep{su2012crowdsourcingAF,ma2022TheEO} and pose estimation~\citep{geng2014headPE} as well as measurements in hardware systems~\citep{zhou2012onTS,zang2019TheIO}.

The vast amount of noisy label learning research has focused more on classification than regression. 
Some notable approaches include regularization~\citep{wang19sce, zhang18nips},
data re-weighting~\citep{ren18l2r,shen19icml}, training procedures~\citep{jiang17icml},
transition matrices~\citep{yao2020dual, xia2020part}, contrastive learning~\citep{zhang2021codim, li2022selective},
refurbishing~\citep{song19b} and sample selection~\citep{lee18cleannet, ostyakov18eccv}.
Particularly, sample selection can be further divided into
exploring the memorability of neural networks~\citep{arpit17memory,zhang17memory}
and delineating samples via the loss magnitude~\citep{wei20jocor}.

To the best of our knowledge, there have been three works that address the noisy label problem for regression with deep learning. 
\citet{castells20} propose a weighted loss correction method based on the small loss assumption.
\citet{garg2020robust} propose an ordinal regression-based loss correction via noise 
transition matrix estimation.
However, they assume that accurate noise rates are known in prior~\citep{patrini17}, which are hard to attain in practice. 
\citet{yao22cmixup} extend MixUp~\citep{zhang18mixup} for regression to interpolate 
the proximal samples in the label space to improve generalization and robustness.
Thanks to its regularizing effect, it can aid the noisy label issue.

In this work, we explore the regression problem with noisy labels, surpassing the scope of previous studies both empirically and methodologically.
For evaluation, we make three notable contributions.
Firstly, recognizing the absence of a standardized benchmark dataset for this task, we take the initiative to curate six balanced real-world datasets. 
These datasets span diverse domains, encompassing age estimation~\citep{niu16afad,lin2021imdbclean}, music production year estimation~\citep{bertin11msd}, and clothing price prediction~\citep{kimura21shift15m}.
Secondly, we conduct a comprehensive empirical benchmark exercise, evaluating the performance of fourteen baselines, which 
are carefully selected from various branches of noisy label research extendable to regression tasks.
Lastly, we introduce a performance measure called Error Residual Ratio (ERR), which accounts for the unique property of regression, where labels exhibit varying degrees of noise severity.

Methodologically, we introduce the ConFrag (Contrastive Fragmentation) framework as a novel approach to address label noise in regression.
It is rooted in one fundamental characteristic of regression: the continuous and ordered correlation between the label and feature space. 
In other words, data points similar in the feature space are likely to have similar labels.
The framework begins by partitioning the dataset into smaller segments, referred to as fragments, and pairs the most distant fragments in the label space to form what we call \textit{contrastive fragment pairs}. 
Training an expert network on these contrastive fragment pairs aids in generalization due to the distinctive feature matching and conversion of closed-set noise into open-set noise, which is less detrimental for learning.
Next, the framework incorporates neighboring relationships by aggregating and reordering the learned features to detect clean samples. 
This is accomplished through the design of Mixture~\citep{jacobs1991MoE} of neighboring fragments.
Furthermore, we enhance our approach with neighborhood jittering regularization, which strengthens the selection process by improving the data coverage of each expert. 
This, in turn, leads to improved agreements among neighboring fragments and serves as an effective tool for mitigating overfitting.
Finally, the contributions of this work can be summarized as follows.
\begin{enumerate}

\item We propose a novel method named ConFrag (Contrastive Fragmentation) for noisy labeled regression. 
    It leverages the inherent orderly relationship within the label and feature space by employing contrastive fragment pairing and constructs a mixture model based on neighborhood agreement. 
    This is further enhanced by our neighborhood jittering regularization.
\item 
    We perform one of the most thorough empirical investigations into noisy labeled regression up to date. 
    We assemble six well-balanced benchmarks using datasets of AFAD~\citep{niu16afad}, IMDB-Clean~\citep{lin2021imdbclean}, IMDB-WIKI~\citep{rothe18imdb}, UTKFace~\citep{zhifei2017utkface}, SHIFT15M~\citep{kimura21shift15m}, and MSD~\citep{bertin11msd}, on which we evaluate fourteen baselines.
    We design a metric termed ERR (Error Residual Ratio), which accounts for the degree of noise severity within the labels, offering a more comprehensive assessment.
 Our experiments affirm the superiority of ConFrag over state-of-the-art noisy label learning baselines.
\end{enumerate}

\section{ConFrag: Contrastive Fragmentation}\label{sec:fragmented_selection}

\begin{figure*}[t]
\begin{center}
\centerline{
\includegraphics[width=\textwidth]{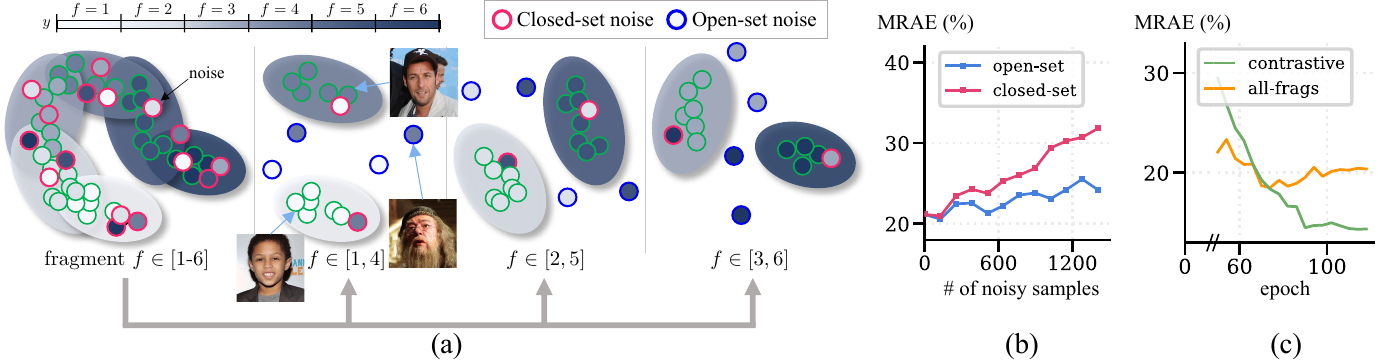}
}
\vskip -0.15in
\caption{
(a) \textbf{An example of t-SNE illustration of contrastive fragment pairing}. 
The data with label noise are grouped into six fragments ($f\in[1\text{-}6]$) and formed into three contrastive pairs ($f\in[1, 4], [2, 5], [3, 6]$). 
Contrastive fragment pairing transforms some of \textcolor[HTML]{E73359}{closed-set noise} (whose ground truth is within the target label set) into \textcolor[HTML]{0000F5}{open-set noise} (whose ground truth is not within the label set).
For example, in the [1,4] figure, label noise whose ground truth fragment is either 1 or 4 is closed-set noise, and the others are open-set noise.
The t-SNE illustration shows that learned features of open-set noises tend to reside outside the feature clusters of the clean samples.
(b) 
The open-set noise is \textit{less harmful} with much lower errors (MRAE) in the downstream regression.
(c) The contrastive pairing ($[1, 4], [2, 5], [3, 6]$) is more effective than using all-fragments together ($[1\text{-}6]$), resulting in much lower MRAE scores.
All experiments are based on IMDB-Clean-B with more details in Appendix~\ref{subsec:contrast_combination}--\ref{subsec:disruptive_anomaly_noise}.
}
\label{fig:fragment_motivation}
\end{center}
\vskip -0.35in
\end{figure*}
In the noisy label regression problem, we are presented with a dataset denoted as $\mathcal{D} = \{\mathcal{X}, Y\}$; in each sample $(x, y)$,  
 $x\in\mathbb{R}^d$ is an input, and $y \in \mathbb{R}$ is the observed label, which can be possibly noisy. We use  $y^\text{gt}$ to denote the groundtruth label. 
The objective of ConFrag is to sample a \textit{clean} subset of the data as $\mathcal{S} \subset \mathcal{D}$. 
By training on $\mathcal{S}$, we aim to enhance the  performance of the regression model.

An overview of our ConFrag framework is shown in Fig.~\ref{fig:framework}(a). 
The framework has the following steps.
We divide the dataset into what we refer to as \textit{contrastive fragment pairs} (\S~\ref{subsec:fragmentation}), which collectively enhance the training of the feature extractors (\S~\ref{subsec:training_feature_extractors}).
We then select clean samples $\mathcal{S}$ from dataset $\mathcal{D}$ based on neighborhood agreements, utilizing a fragment-based mixture model (\S~\ref{subsec:mixture_of_contrasing_fragments}). 
A regression model is trained on the clean samples $\mathcal{S}$.
We also propose neighborhood jittering as a regularizer for further improved training (\S~\ref{sec:jittering}).
ConFrag is noise rate-agnostic unlike prior methods as it operates without knowing a pre-defined noise rate.


\subsection{Contrastive Fragment Pairing}\label{subsec:fragmentation}

\begin{wrapfigure}{r}{0.45\textwidth}
\begin{center}
\vskip -0.1in
\centerline{\includegraphics[width=0.5\columnwidth]{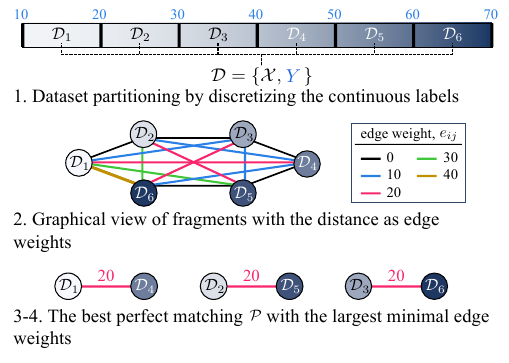}}
\caption{The contrastive fragment pairing algorithm.} 
\label{fig:contrastive_fragmentation}
\end{center}
\vskip -0.3in
\end{wrapfigure}

In order to sample a \textit{clean} data subset $\mathcal{S} \subset \mathcal{D}$, 
we need to learn a robust feature that can distinguish clean samples from noisy ones.
As one theoretical result in \citep{zhang2023}, the cross-entropy loss used in classification is better for learning high-entropy feature representation than the mean squared loss in regression (see Appendix~\ref{subsec:classification_vs_regression} for details). 
Based on this, we start by discretizing the label space into $F$ continuous fragments, transforming the original regression problem into the multi-class classification one.
This transformation harnesses an inherent property of regression: data points with similar labels are also represented with closely related features, as acknowledged in prior studies \citep{gong22rank,yang21ldsfds,yao22cmixup}.

However, instead of training a single feature extractor on the multi-class classification with $F$ classes, we construct $F/2$ \textit{maximally contrasting fragment pairs} and train a smaller expert feature extractor for each pair.
The procedure of contrastive fragment pairing is detailed below with an illustration in Fig.~\ref{fig:contrastive_fragmentation}:

\begin{enumerate}[leftmargin=*]
\item Divide the range of continuous labels $Y$ into $F$ even number of equal-length fragments.
  This allows to divide the dataset $\mathcal{D}$ into $F$ disjoint subsets: $\mathcal{D} = \{\mathcal{D}_1, ... ,\mathcal{D}_F \}$, where 
    each $\mathcal{D}_i$ contains the data samples whose $y$ values are in the $i$-th fragment label range. 
\item Construct a complete graph $g= \{\mathcal{D}, E\}$, where each vertex is a  fragment $\mathcal{D}_i$, and each edge weight $e_{ij}$ is the distance in the label space between the closest samples of the fragments $(\mathcal{D}_i,\mathcal{D}_j)$. 
\item Compute all possible \textit{perfect matchings}~\citep{monfared16pm, gibbons85pm}, where every vertex of a graph is incident to exactly one edge in the graph.  
\item Find the perfect matching with the largest minimal edge weight: 
        $\mathcal{P} = \argmax_{\bar{g}\in \mathcal{G}}\Bigl(\min\nu(\bar{g})\Bigr)$,
    where each $\bar{g}$ is a perfect matching (graph), and  $\nu(\bar{g})$ is the set of edge weights in $\bar{g}$.
Finally, $\mathcal{P}=\{ (\mathcal{D}_i, \mathcal{D}_j), \ldots ,(\mathcal{D}_k, \mathcal{D}_l)\}$ constitutes the \textit{maximally contrasting pairs} of fragments. 
\end{enumerate}

\textbf{Motivation behind contrastive fragment pairing. }
Formulating the multi-class classification problem into $F/2$ binary classification problems via contrastive fragment pairing has the following advantages.
Firstly, since the distance between fragments in each contrastive fragment pair is large, the feature extractor trained on each contrastive pair can generalize better \citep{shawe1998robust, gronlund2019margin, gronlund2020near}.
Fig.~\ref{fig:fragment_motivation}(c) shows the generalization abilities of the expert feature extractors trained on contrastive fragment pairs compared to the single feature extractor trained on all fragments.
When using a single feature extractor on all fragments (all-frags), the samples selected by the feature extractor tend to become more noisy as the feature extractor overfits, causing the regressor to perform worse over time.
On the other hand, when using multiple feature extractors trained on contrastive pairs, the performance of the regression model consistently improves, indicating that the learned features are more robust and the selected samples are cleaner.
The large distance between fragments also explains why contrastive fragment pairing is superior to other fragment pairings, as shown in \S~\ref{sec:discussion}.
The analysis of the prediction depth \citep{baldock21nips} in Appendix~\ref{subsec:prediction_depth_analysis} supports the claim, as it shows that the binary classification on contrastive fragment pairs results in lower prediction depth, leading to better generalization.

Secondly, the contrastive fragment pairing transforms some of \textit{closed-set} label noise (whose ground truth is within the label set) into \textit{open-set} label noise (whose ground truth is not within the label set), as shown in Fig.~\ref{fig:fragment_motivation}(a).
Previous works \citep{wei2021open, wan2024unlocking} observe that the open-set noise is less harmful than the closed-set noise and may even benefit generalization and robustness against inherent noisy labels.
Indeed, in our experiments, we found similar observations where injecting open-set label noise is less harmful than closed-set one, as shown in Fig.~\ref{fig:fragment_motivation}(b).

The t-SNE visualization in Fig.~\ref{fig:fragment_motivation}(a) also supports this observation.
Let $f$ and $f^\text{gt}$ be fragments that the observed label $y$ and the groundtruth label $y^\text{gt}$ respectively belong to.
Prior to contrastive fragment pairing, all of the noisy labeled data ($f\neq f^\text{gt}$) are \textcolor[HTML]{E73359}{closed-set noise} as their ground truth fragment ids are within the label set ($f^\text{gt}\in[1\text{-}6]$) and their features 
are located in the feature spaces of incorrect classes within the group.
After contrastive fragment pairing, much of these noisy labeled data is transformed to \textcolor[HTML]{0000F5}{open-set noise} ($f^\text{gt}\notin[1,4]$ while $f\in[1,4]$ in case of fragment pair $[1, 4]$), 
and their learned features tend to reside outside the feature clusters of the clean samples, thus mitigating the adverse effects of the noise.


\subsection{Training  Feature Extractors for Contrastive Pairs}
\label{subsec:training_feature_extractors}

Once we obtain the contrastive fragment pairs $\mathcal{P}$, we train $F/2$ number of expert feature extractors on binary classification $p(y|x;\theta_{i,j})$ with its respective contrastive pair $(\mathcal{D}_i, \mathcal{D}_j) \in \mathcal{P}$, where $\theta_{i,j}$ denotes the parameter of an expert.
That is, it is trained to predict whether a data $x$ is in $\mathcal{D}_i$ or $\mathcal{D}_j$.
Later, the feature extractors play a crucial role in determining whether a sample $(x, y)$ is clean.

\begin{figure}[t]
\begin{center}
\includegraphics[width=\textwidth]{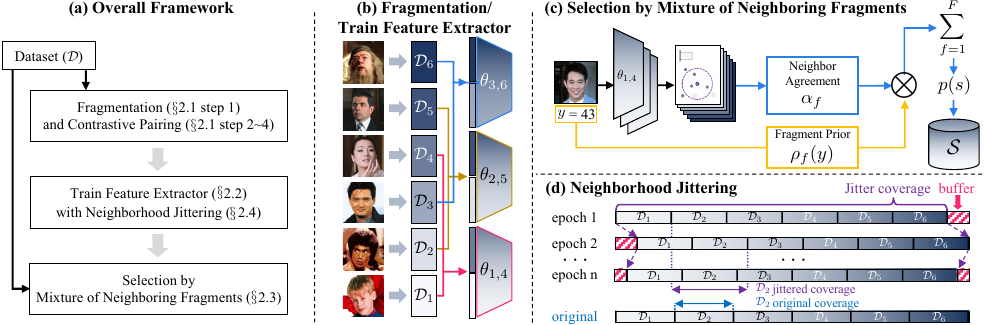}
\caption{\textbf{Contrastive Fragmentation framework.}
(a) The overall sequential process of our framework.
(b) Shows the fragmentation of the continuous label space to obtain \textit{contrasting fragment pairs} (\S~\ref{subsec:fragmentation}) and train feature extractors on them.
(c) Sample Selection by Mixture of Neighboring Fragments obtains the selection probability in both prediction and representation perspectives (\S~\ref{subsec:mixture_of_contrasing_fragments}).
(d) Illustration of Neighborhood Jittering (\S~\ref{sec:jittering}).
}
\label{fig:framework}
\end{center}
\vskip -0.2in
\end{figure}

\subsection{Mixture of Neighboring Fragments}
\label{subsec:mixture_of_contrasing_fragments}

With the learned expert feature extractors, the next step is to perform sample selection.
Given a sample $(x, y)$, let $f$ be a fragment close to $y$ and $f^+$ be its contrasting pair.
Intuitively, we consider a sample clean if the expert trained on $(\mathcal{D}_f, \mathcal{D}_{f^+})$ strongly predicts that $x$ belongs to a fragment $f$.
However, since the expert feature extractor is a binary classifier only trained using a contrasting pair of fragments, we utilize all experts' opinions to obtain a more robust prediction. 
Specifically, we deem a sample as clean if the experts exhibit a consensus response (\textbf{Neighborhood Agreement}) for fragments close to $y$ (\textbf{Fragment Prior}).

Based on this intuition, we formulate Mixture of Experts (MoE) \citep{jacobs1991MoE} model, where the sampling probability of a datapoint $(x, y)$ is defined as
\begin{align}\label{eq:mcf}
    p(s|x,y, \mathcal{D}_{1 \ldots F};\Theta) &= \sum_{f}^{F} \rho_f(y)\alpha_f(x; \mathcal{D}_{1 \ldots F}, \Theta),
\end{align}
where $\Theta$ denotes parameters of all  feature extractors, $\rho_f$ is the \textit{fragment prior} (mixture weight), and $\alpha_f$ is the \textit{neighborhood agreement} (a binary vote  of whether $x$ belongs to the fragment $f$).
Based on the intuition above, $\rho_f(y)$ is large when the fragment $f$ is close to $y$, and $\alpha_f(x) \in \{0, 1\}$ is $1$ if $x$ is likely to belong to the fragment $f$.

\textbf{Fragment Prior.}
For a sample $(x, y)$, we compute the prior $\rho_f(y)$ of a fragment $f$,
using a softmax weighting of each fragment $f$ with respect to its relative distance to $y$:
\begin{align}\label{eq:np}
    \rho_f(y)=\frac{\exp(g_f(y))}{\sum_{f'}^F \exp(g_{f'}(y))}, 
\end{align}
where $g_f(y)=\text{range}(Y)/ (|y - \bar{Y}_f|)$, $\text{range}(Y)=\max(Y) - \min(Y)$ is the label range, and $\bar{Y}_f$ is the mean label value of fragment $f$.
Since $\text{range}(Y)$ is a constant for a given dataset, $g_f(y)$ rapidly decreases when the mean value of fragment $f$ is far from $y$ in the continuous label space.
From the MoE perspective, the fragment prior can be regarded as soft gating that depends on $y$.

\textbf{Neighborhood Agreement.}
%
Given a sample $(x, y)$ and a fragment $f$, we need to determine whether $x$ belongs to $f$.
The simplest approach is to use the expert trained using $(\mathcal{D}_f, \mathcal{D}_{f^+})$ to classify whether $x$ belongs to $f$ or $f^+$, where $f^+$ is the contrasting fragment of $f$.
Based on the classification output $h(x; \theta_{f, f^+}) \in \{f, f^+\}$, we define self-agreement as:
\begin{align}
   \alpha^\text{self}_f = 
       [h(x; \theta_{f, f^+}) = f] 
\label{eq:self_agreement}
\end{align}
where $[A]$ is the Iverson bracket outputting $1$ if $A$ is true, and 0 otherwise.
Since training with noisy labels often results in suboptimal calibration \citep{bae22icml, zong2024dirichlet}, we use discrete classification output for $\alpha^\text{self}_f$ rather than continuous probabilistic one.

Since the expert $\theta_{f, f^+}$ is only trained to discriminate between $f$ and its contrasting fragment $f^+$, it is better to utilize other experts to obtain a more robust prediction.
For example, consider contrastive fragment pairs $\{(1,4), (2,5), (3,6)\}$ as in Fig.~\ref{fig:contrastive_fragmentation}.
If $x$ is more likely to belong to fragment $2$ than $5$, then it should be more likely to belong to $1$ than $4$ and $3$ than $6$.
Thus, we consider agreement of neighboring fragments $f_L$ (left) and $f_R$ (right) to obtain neighborhood agreement $\alpha_f(x; \mathcal{D}_{1 \ldots F}, \Theta)$:
\begin{align}
  \alpha_f(x; \mathcal{D}_{1 \ldots F}, \Theta) = \alpha^\text{self}_f \cdot \alpha^\text{ngb}_f, \hspace{3pt} \text{ where} 
    \ \ \alpha^\text{ngb}_f = \left[ \alpha^\text{self}_{f_L} \lor \alpha^\text{self}_{f_R} \right]. \label{eq:na_final}    
\end{align}
Intuitively, $\alpha_f$ is $1$ if the fragment $f$ is more likely for $x$ than $f^+$ ($\alpha^\text{self}_f = 1$) and either $f$'s left or right fragment is more likely for $x$ than its respective contrasting fragment ($\alpha^\text{ngb}_f = 1$).

In practice, we implement two variants of the agreements in Eq.(\ref{eq:self_agreement}--\ref{eq:na_final}) using the feature extractor's binary classifier and a $K$-nearest neighbor classifier on the learned feature space.
These two classifiers respectively consider \textit{predictive} and \textit{representational} aspects of the expert feature extractor and effectively work as an ensemble, as shown in Appendix~\ref{subsec:ablation}.
As a result, we compute two versions of sample probability in Eq.(\ref{eq:mcf}), and use the union of the sampled \textit{clean} dataset $\mathcal{S}$ for training of the regression model. 
Algorithm~\ref{alg:fragmented_selection} in Appendix summarizes the overall procedure.

\subsection{Neighborhood Jittering}\label{sec:jittering}

\begin{figure*}[t]
\begin{center}
\centerline{\includegraphics[width=1.0\textwidth]{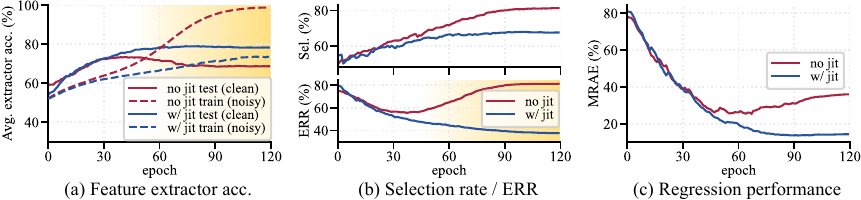}}
\vskip -0.1in
\caption{\textbf{Jittering analysis.}
(a) When trained without jittering, feature extractors easily overfit the noisy training data (yellow-shaded region), while jittering-regularized feature extractors robustly learn from the noisy training data. 
(b) Overfitted feature extractors (yellow-shaded region) on noisy samples increase their likelihood, leading to a higher selection rate and ERR. 
It exhibits nearly twice higher ERR (a lower value is better).
(c) Most importantly, jittering regularization improves performance in regression.
}
\label{fig:jitter_analysis}
\end{center}
\vskip -0.2in
\end{figure*}

A potential limitation of mixture models is that the individual expert feature extractor may not fully benefit from the full dataset as they model their own disjoint subsets~\citep{dukler23}.
Our neighborhood jittering mitigates this limitation as a robust regularizer that expands the effective coverage of each contrastive fragment pair during learning.
The process is visualized in Fig.~\ref{fig:framework}(d).

We bound the ratio of the jittering buffer range  within $[0, \frac{1}{2(F-1)}]$, where $F$ is the fragment number.
For every epoch, we shift the label coverage of each fragment by randomly sampling the value in this range. 
Jittering leads to a partially overlapping mixture model~\citep{heller07, hinton02poe} as some data belong to multiple, neighboring fragments and thus the effective coverage per each expert is expanded.
Such regularization inhibits feature extractors from overfitting to potentially noisy samples and promotes learning of more robust features, even those that can be generalizable to overlapping parts of neighboring fragments.

Fig.~\ref{fig:jitter_analysis}(a) shows that with jittering, the feature extractor exhibits higher accuracy on the clean test data due to its regularization effect. 
In the sample selection stage (Fig.~\ref{fig:jitter_analysis}(b)), 
the feature extractor trained without jittering easily overfits the noise, resulting in over-selection and higher ERR (\S~\ref{subsec:evaluation_metrics}). 
In contrast, the jittered feature extractor achieves a relatively low selection rate with halved ERR, indicating that the noisier samples are filtered out.
Better sample selection due to jittering subsequently leads to significantly better performance in regression (Fig.~\ref{fig:jitter_analysis}(c)).
In Appendix~\ref{subsec:jitter_comparison}, we compare neighborhood jittering to other regularizations, demonstrating its efficacy.



\section{Related Works}\label{sec:related_works}
\vskip -0.05in
We review prior works on learning with noisy labels and defer a comprehensive survey to Appendix \ref{subsec:related_work}. 
We organize them into those utilizing prediction, representation, and combination of the two.

\textbf{Prediction-based Methods}.
This approach has been the focus of much existing research and covers a wide array of topics: (i) 
the small loss selection by exploring the pattern of memorization in neural networks~\citep{han18coteaching, arazo19}, 
(ii) relying on the consistency of predictions to select or refurbish the samples~\citep{liu2020early,huang2020self},
(iii) estimating the noise distribution~\citep{patrini17,hendrycks18nips},
(iv) introducing auxiliary parameters or labels~\citep{pleiss20aum,hu20rdiaux},
(v) using unlabeled data with semi-supervised learning~\citep{li2020dividemix,bai2021understanding,karim2022unicon}, 
and (vi) designing a noise-robust loss function~\citep{menon20phuber,wang19sce}.

\textbf{Representation-based Methods}.
This approach has seen a recent surge in interest, including
(i) clustering based selection~\citep{mirzasoleiman20crust,wu20topo},
(ii) feature eigendecomposition filtering~\citep{kim21fine}, 
(iii) using neighbor information to sample and refurbish with clean validation~\citep{li22neighbor, gao16knn},
and (iv) generative models of features for sampling~\citep{kmlee19}.

\textbf{Combination}.
Some works have also studied the combination of representation and prediction spaces.
\citet{wang22spr} formulate a penalized regression between the network features and the labels for selection,
and \citet{ma18d2l} use intrinsic dimensionality and consistent predictions to refurbish.
Other important approaches include (i) regularization via MixUp~\citep{zhang18mixup} along with its regression version~\citep{yao22cmixup},
(ii) model-based methods that discourage large parameter shifts~\citep{hu20rdiaux}, and (iii) importance discrimination of parameter updates~\citep{xia21cdr}.

The majority of previous works have studied noisy labels for classification. Hence, a large portion of these works may not be directly applicable to regression tasks due to the restricted usage of class-wise information. 
In \S~\ref{sec:experiments}, we empirically compare our method with some of these works that are expandable to the regression task with some or minor technical adaptation.

\begin{table*}[th!]
    \caption{\textbf{Comparison of Mean Relative Absolute Error (\%)} over the noise-free trained model on the AFAD-B, IMDB-Clean-B, IMDB-WIKI-B, SHIFT15M-B, and MSD-B datasets.
    Lower is better. A negative value indicates it performs even better than the noise-free model.
    The results are the mean of three random seed experiments.
    The best and the second best methods are respectively marked in \textcolor{red}{red} and \textcolor{blue}{blue}.
    CNLCU-S/H, Co-Selfie, and Co-ConFrag use dual networks to teach each other as done in \citet{han18coteaching}.
    SPR~\citep{wang22spr} fails to run for SHIFT15M-B due to excessive memory usage.}
    \begin{center}
    \begin{small}
    \setlength{\tabcolsep}{2.0pt}
    \begin{tabular}{lccccccccccccc}
        \toprule
        &\multicolumn{6}{c}{AFAD-B}       &\multicolumn{6}{c}{IMDB-Clean-B}& IMDB-WIKI-B
        \\\cmidrule(lr){2-7}\cmidrule(lr){8-14}
        &\multicolumn{4}{c}{symmetric}    &\multicolumn{2}{c}{Gaussian} &\multicolumn{4}{c}{symmetric} &\multicolumn{2}{c}{Gaussian} & real noise
        \\\cmidrule(lr){2-5}\cmidrule(lr){6-7}\cmidrule(lr){8-11}\cmidrule(lr){12-13}\cmidrule(lr){14-14}
        noise rate  & 20 & 40 & 60 & 80 & 30 & 50 & 20 & 40 & 60 & 80 & 30 & 50 & - \\
        \midrule
        Vanilla & 9.37  & 20.27 & 30.65 & 43.09 & 28.77 & 39.03 & 16.18 & 32.05 & 53.13 & 76.35 & 26.89 & 50.28 & 0 \\
        \specialrule{0.1pt}{1pt}{1pt}
        CNLCU-S   & 10.98 & 20.44 & 32.44 & 41.99 & 30.60 & 40.66 & 51.40 & 66.62 & 82.83 & 85.65 & 83.39 & 82.10 & 21.54 \\
        CNLCU-H   &  4.63 & 16.32 & 36.01 & 44.71 & 35.68 & 43.64 & 6.84 & 31.16 & 63.08 & 82.65 & 46.53 & 65.24 & -2.93 \\
        Sigua     &  5.96 & 21.09 & 43.33 & 49.71 & 42.52 & 46.19 & 9.82 & 46.17 & 77.59 & 85.62 & 60.97 & 77.42 & 1.96 \\
        SPR       &  9.74 & 18.85 & 30.43 & 43.25 & 28.50 & 39.69 & 14.47 & 32.44 & 54.88 & 79.37 & 25.67 & 51.05 & -0.93 \\
        BMM       &  5.60 & 15.00 & 39.15 & 46.41 & 30.96 & 44.00 & 8.85 & 21.54 & 55.57 & 80.40 & 24.33 & 57.21 & 17.88 \\
        DY-S      &  6.87 & 15.56 & 32.24 & 45.72 & 24.40 & 43.41 & 10.42 & 21.90 & 49.94 & 78.16 & 24.70 & 44.56 & -3.41 \\
        C-Mixup   & \textcolor{blue}{2.74} & 14.80 & 27.17 & 41.95 & 24.28 & 36.91 & 8.82 & 27.74 & 50.87 & 76.79 & 21.92 & 47.04 & \textcolor{blue}{-5.26} \\
        RDI       & 10.64 & 21.80 & 39.32 & 47.07 & 37.33 & 44.41 & 16.35 & 29.33 & 55.91 & 79.92 & 25.69 & 51.35 & 1.06 \\
        CDR       & 10.26 & 18.71 & 32.27 & 43.38 & 29.74 & 39.21 & 17.47 & 32.19 & 54.75 & 75.45 & 28.46 & 51.73 & -0.39 \\
        D2L       &  9.43 & 20.75 & 31.25 & 44.50 & 28.86 & 40.10 & 16.94 & 33.85 & 55.54 & 76.28 & 29.30 & 52.44 & -0.66 \\
        AUX       &  6.15 & 19.01 & 31.16 & 42.83 & 28.28 & 39.05 & 12.58 & 28.82 & 52.33 & 76.75 & 23.27 & 49.42 & -3.67 \\
        Selfie    & 16.91 & 25.02 & 44.18 & 47.78 & 46.02 & 50.73 & 27.43 & 53.74 & 79.38 & 84.00 & 60.68 & 78.03 & 14.00 \\
        Co-Selfie & 14.61 & 22.95 & 39.79 & 47.72 & 41.05 & 53.00 & 23.52 & 50.07 & 67.42 & 84.25 & 52.44 & 74.73 & -0.44 \\
        Superloss &  7.36 & 18.24 & 29.78 & 44.26 & 27.59 & 42.96 & 23.38 & 45.41 & 67.11 & 80.85 & 53.88 & 63.33 & -3.58 \\
        \specialrule{0.7pt}{1pt}{1pt}
        \textbf{ConFrag}  & \textcolor{blue}{2.74} & \textcolor{blue}{8.16} & \textcolor{red}{15.91} & \textcolor{blue}{34.42} & \textcolor{blue}{17.49} & \textcolor{red}{27.31} & \textcolor{blue}{5.08} & \textcolor{blue}{12.64} & \textcolor{red}{27.26} & \textcolor{red}{61.24} & \textcolor{blue}{15.70} & \textcolor{red}{33.36} & -3.06 \\
        \textbf{Co-ConFrag}  & \textcolor{red}{0.54} & \textcolor{red}{7.25} & \textcolor{blue}{16.65} & \textcolor{red}{33.93} & \textcolor{red}{17.43} & \textcolor{blue}{28.26} & \textcolor{red}{1.50} & \textcolor{red}{9.45} & \textcolor{blue}{28.44} & \textcolor{blue}{61.36} & \textcolor{red}{14.87} & \textcolor{blue}{35.88} & \textcolor{red}{-8.86} \\
        \bottomrule
        \\
    \end{tabular}
    \begin{tabular}{lcccccccccccc}
        \toprule
        &\multicolumn{6}{c}{SHIFT15M-B}         &\multicolumn{6}{c}{MSD-B}
        \\\cmidrule(lr){2-7}\cmidrule(lr){8-13}
        &\multicolumn{4}{c}{symmetric}    &\multicolumn{2}{c}{Gaussian} &\multicolumn{4}{c}{symmetric} &\multicolumn{2}{c}{Gaussian}
        \\\cmidrule(lr){2-5}\cmidrule(lr){6-7}\cmidrule(lr){8-11}\cmidrule(lr){12-13} 
        noise rate & 20 & 40 & 60 & 80 & 30 & 50 & 20 & 40 & 60 & 80 & 30 & 50 \\
        \midrule
        Vanilla            & 9.11 & 17.96 & 27.02 & 36.34 & 6.54 & 15.16 & 8.23 & 18.43 & 31.67 & 45.85 & 6.96 & 15.74 \\
        \specialrule{0.1pt}{1pt}{1pt}
        CNLCU-S & 12.98 & 19.42 & 24.31 & 34.47 & 15.33 & 20.90 & 0.13 & 6.04 & 21.52 & 46.01 & 4.75 & 12.51 \\
        CNLCU-H & 6.26 & 12.84 & 20.04 & 36.03 & 8.88 & 15.65 & 0.27 & 4.98 & 10.32 & 29.83 & 5.11 & 9.22 \\
        Sigua & 6.94 & 14.09 & 26.08 & 37.03 & 10.32 & 17.44 & 1.29 & 7.19 & 17.35 & 50.87 & 6.80 & 12.38 \\
        SPR &-&-&-&-&-&-& 7.07 & 18.19 & 33.39 & 45.61 & 5.01 & 15.36 \\
        BMM & 6.96 & 12.42 & 18.64 & 26.79 & 7.58 & 13.13 & 3.32 & 10.30 & 23.40 & 43.56 & 5.29 & 11.85 \\
        DY-S & 7.11 & 11.94 & 18.85 & 29.04 & 6.90 & 13.50 & 3.39 & 8.06 & 18.65 & 35.24 & 4.77 & 9.83 \\
        C-Mixup & 9.47 & 16.15 & 24.08 & 34.17 & 5.88 & 14.51 & 3.75 & 13.13 & 26.73 & 40.90 & 2.96 & 10.97 \\
        RDI & 9.91 & 17.92 & 26.63 & 36.29 & 7.08 & 15.18 & 21.04 & 30.09 & 38.78 & 49.49 & 19.19 & 27.88 \\
        CDR & 9.52 & 17.78 & 26.97 & 35.97 & 7.14 & 15.17 & 7.83 & 17.86 & 32.83 & 45.91 & 6.73 & 16.92 \\
        D2L & 9.25 & 18.03 & 26.55 & 36.23 & 6.34 & 15.60 & 7.13 & 19.96 & 32.47 & 46.64 & 5.51 & 15.54 \\
        AUX & 7.74 & 16.95 & 26.61 & 36.47 & 4.92 & 14.40 & 6.12 & 18.18 & 31.09 & 45.70 & 5.21 & 15.45 \\
        Selfie & 4.84 & 10.22 & 22.28 & 38.15 & 5.51 & 11.58 & 1.43 & 8.40 & 20.24 & 45.87 & 14.37 & 24.13 \\
        Co-Selfie & 11.53 & 16.43 & 32.08 & 39.32 & 13.45 & 22.33 & \textcolor{blue}{-0.38} & \textcolor{blue}{4.41} & \textcolor{red}{8.32} & 35.47 & 6.78 & 13.15 \\
        Superloss & 5.44 & 12.26 & 23.23 & 35.24 & 5.60 & 13.28 & -0.15 & 10.68 & 23.15 & 45.55 & 4.35 & 16.36 \\
        \specialrule{0.7pt}{1pt}{1pt}
        \textbf{ConFrag} & \textcolor{blue}{2.46} & \textcolor{blue}{6.18} & \textcolor{red}{10.68} & \textcolor{blue}{19.04} & \textcolor{blue}{3.66} & \textcolor{red}{8.09} & 0.57 & 4.94 & 11.22 & \textcolor{blue}{23.41} & 2.39 & \textcolor{blue}{6.49} \\
        \textbf{Co-ConFrag} & \textcolor{red}{0.85} & \textcolor{red}{5.52} & \textcolor{blue}{10.80} & \textcolor{red}{18.83} & \textcolor{red}{3.03} & \textcolor{blue}{8.70} & \textcolor{red}{-0.65} & \textcolor{red}{2.98} & \textcolor{blue}{8.66} & \textcolor{red}{20.53} & \textcolor{red}{1.73} & \textcolor{red}{6.00} \\
        \bottomrule
    \end{tabular}
    \end{small}
    \end{center}
    \label{tab:main_mrae}
    \vskip -0.15in
\end{table*}

\section{Experiments}\label{sec:experiments}
\vskip -0.05in
We compare ConFrag with fourteen strong baselines adapted for noisy label regression.
Due to the scarcity of benchmark datasets, we update existing datasets for the study of noisy labels.

\subsection{Settings}
\label{subsec:experiment_settings}
\textbf{Curation of Benchmark Datasets.}
We create six benchmark datasets for noisy labeled regression to encompass a sufficient quantity of balanced data, span multiple domains, and present a meaningful level of complexity. 
(i) \textit{Age Prediction} from an image is a well-studied regression problem~\citep{li19bridge, shin2022moving,lim20order}. 
To address this domain, we acquire four datasets of \textbf{AFAD}~\citep{niu16afad}, \textbf{IMDB-Clean}~\citep{lin2021imdbclean}, \textbf{IMDB-WIKI}~\citep{rothe18imdb}, and \textbf{UTKFace}~\citep{zhifei2017utkface}.
Notably, IMDB-WIKI contains real-world label noise stemming from the automatic web crawling process~\citep{lin2021imdbclean}.
We use a ResNet-50 backbone for all datasets. 
(ii) \textit{Commodity Price Prediction} is a vital real-world task~\citep{wen2021fashion}.
We opt for the \textbf{SHIFT15M} dataset~\citep{kimura21shift15m} due to the diversity and scale of this domain.
This dataset is provided as the penultimate feature of the ImageNet pre-trained VGG-16 model. 
Consequently, we use a three-layer MLP architecture for all experiments \citep{papadopoulos22fashion,kimura21shift15m}.
(iii) \textit{Music Production Year Estimation} uses the tabular \textbf{MSD} dataset~\citep{bertin11msd}. 
This dataset is identified as one of the most intricate and challenging datasets, based on the test R2 score \citep{grinsztajn22nips}.
We adopt a tabular ResNet proposed by \citet{gorishniy21nips}.
The suffix ``-B'' is appended to the dataset name (\eg AFAD-B) to indicate that it is a curated version of the original dataset.
To focus on the noisy label problem, we take measures to balance the datasets as elaborated in Appendix~\ref{subsec:dataset_curation}.


\textbf{Experimental Design.}
For all datasets except for IMDB-WIKI-B which contains real-world label noise, 
we inject symmetric and Gaussian noise into the labels, as done in prior literature~\citep{yao22cmixup, yi19pencil,wei20jocor}.
These types of noise can simulate a low-cost (human-free) controlled setting. 
Symmetric noise mimics randomness such as Web crawling or annotator errors, and
Gaussian noise is often used for modeling the regression label noise. 
While \citet{yao22cmixup} inject a \textit{fixed} 30\% standard deviated Gaussian noise for \textit{every label},
we make it more realistic by \textit{randomizing} the standard deviation up to 30\% or 50\% of the domain's range.
For our ConFrag experiments, we fix the fragment number ($F$) as four. 
See Appendix~\ref{subsec:training_details} for further training details.

\textbf{Baselines.} There are many existing methods of noisy labeled learning for classification. 
We assess fourteen baselines from the three branches that are naturally adaptable to regression with minor or no updates. 
(i) Small loss selection: CNLCU-S,H~\citep{xia22}, Sigua~\citep{han20sigua}, SPR~\citep{wang22spr}, BMM~\citep{arazo19}, DY-S~\citep{arazo19}, SuperLoss~\citep{castells20}.
(ii) Regularization: C-mixup~\citep{yao22cmixup}, RDI~\citep{hu20rdiaux}, CDR~\citep{xia21cdr}, D2L~\citep{ma18d2l}.
(iii) Refurbishment: AUX~\citep{hu20rdiaux}, Selfie~\citep{song19b}, Co-Selfie~\citep{song19b}.
Appendix~\ref{subsec:baselines} comprehensively details these baselines.

\begin{figure*}[t]
\begin{center}
\centerline{\includegraphics[width=\textwidth]{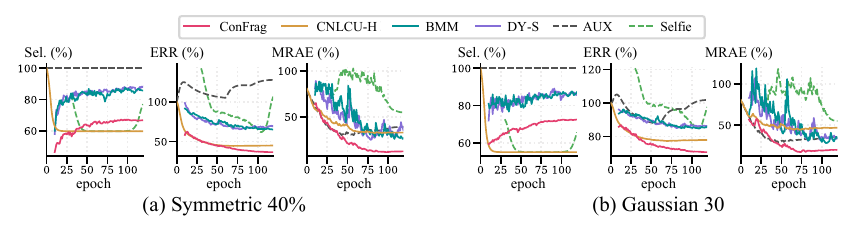}}
\vskip -0.1in
\caption{\textbf{Selection/ERR/MRAE comparison}
between ConFrag and strong baselines of CNLCU-H, BMM, DY-S, AUX and Selfie on IMDB-Clean-B. We exclude the performance during the warm-up. 
}
\label{fig:hse}
\end{center}
\vskip -0.25in
\end{figure*}

\subsection{Evaluation Metrics}\label{subsec:evaluation_metrics}
We mainly report the Mean Relative Absolute Error (MRAE) following prior works. 
The MRAE is computed as $(e/\rho)-1$, where $e$ is the model's Mean Absolute Error (MAE) performance under varying conditions (noise type, severity) and $\rho$ is the noise-free model's MAE. 
We express MRAEs in percentage for better comprehensibility. 
The traditional MAE values are also reported in Appendix~\ref{subsec:mae}.
In addition, we report the Selection rate (a.k.a prevalence), which is a metric often seen in noisy labeled classification to quantify the coverage of the total dataset, $|\mathcal{S}|/|\mathcal{D}|$ where $\mathcal{S}$ and $\mathcal{D}$ are the selected and total set, respectively. 

\textbf{Error Residual Ratio}. 
To better assess selection and refurbishment approaches, we introduce a new metric termed Error Residual Ratio (ERR).
Unlike classification, noisy labels in regression can show the diverse severity of the noise present in each label $y$ (\ie various degrees of deviation from the ground truth $y^\text{gt}$). 
This cannot be addressed when using conventional metrics, which are primarily designed for classification and tend to treat all instances of noise as equally severe. 
Our proposed ERR considers the varying severity of noise and is defined as
\begin{gather}\label{eq:ERR}
   \text{ERR} = \frac{1/|\mathcal{C}|\sum_c^{|\mathcal{C}|} |y_c - y^\text{gt}_c|}{1/|\mathcal{D}|\sum_d^{|\mathcal{D}|} |{y}^\text{ }_d - y^\text{gt}_d|},
\end{gather}
where $\mathcal{C}$ is a set of cleaned (selected or refurbished) samples. 
The numerator is the average cleaned error that serves as an indicator of the precision of the cleaned data, while the denominator is the average dataset error that normalizes it for standardized assessment.
The ERR, along with the selection rate and regression metrics (\eg MSE, MRAE), provides a deeper insight into the model performance.
Ideally, a method with a high selection rate coupled with low ERR and regression error can be deemed as closer to the upper bound.

\subsection{Results and Discussion}\label{sec:discussion}
\textbf{Overall performance.}
Table~\ref{tab:main_mrae} compares the MRAE values to 
the noise-free trained model between ConFrag and the baselines.
We evaluate six types of noise: four symmetric and two random Gaussian noises. 
ConFrag and Co-ConFrag achieve the strongest performance in all experiments compared to the fourteen baselines.
Notably, Co-ConFrag mixes co-teaching during the regression learning phase by assuming that $\mathcal{S}$ still contains 25\% noise.
The results on UTKFace-B dataset can be found in Appendix~\ref{subsec:utkface}.

\textbf{Selection/ERR/MRAE comparison.} 
Fig.~\ref{fig:hse} compares ConFrag to five selection and refurbishment baselines of CNLCU-H, BMM, DY-S, AUX, Selfie on IMDB-Clean-B using the selection rate, ERR, and MRAE.
Ideally, a model should attain a high selection rate and a low ERR.
It is worth noting that the relative importance of ERR and selection rate may vary depending on the dataset and the task. 
ConFrag achieves the lowest ERR while maintaining above-average selection rates, resulting in the best MRAE.
Appendix~\ref{subsec:ERR} includes comparison results for all noise types with more baselines.

\begin{figure}[t]
    \begin{center}
    \centerline{\includegraphics[width=0.95\textwidth]{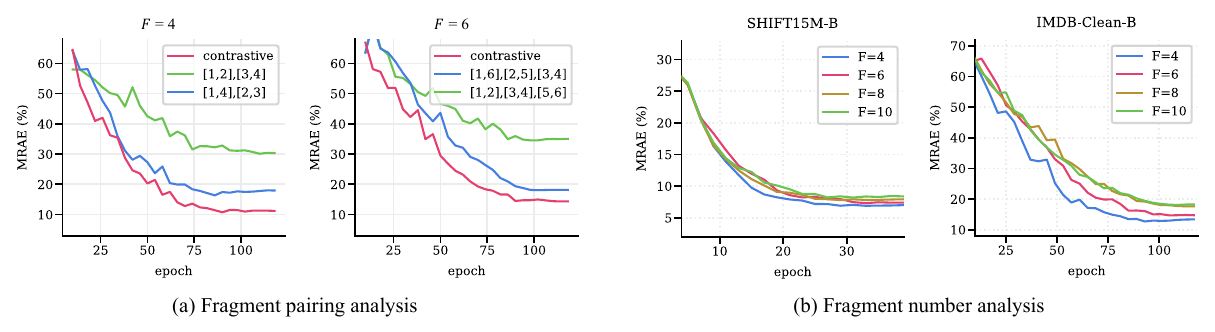}}
    \vskip -0.1in
    \caption{
    \textbf{Analysis} with 40\% symmetric noise. (a) Comparison between the proposed contrastive pairing  and other pairings on IMDB-Clean-B.
    (b) Comparison between fragment numbers on SHIFT15M-B and IMDB-Clean-B.
    }
    \vskip -0.3in
    \label{fig:discussion_main}
    \end{center}
    \end{figure}

\textbf{Fragment pairing.}
Fig.~\ref{fig:discussion_main}(a) compares contrastive pairing to alternative pairings 
using MRAE as a metric. 
The contrastive fragment pairing demonstrates superior performance to other pairing methods.
Notably, the performance is poorest when both the average and minimum distance between fragments are smallest ($[1,2], [3,4]$ when $F=4$, $[1,2],[3,4],[5,6]$ when $F=6$).
While the pairings of $[1,4], [2,3]$ and $[1,6],[2,5],[3,4]$ have the same average distance between fragments as the contrastive pairings, their minimum distances between fragments are smaller, resulting in poorer performances than contrastive pairings.
This result shows the effectiveness of contrastive fragment pairing for selecting clean samples.
See Appendix~\ref{subsec:contrast_combination} for more details. 

\textbf{Fragment number.}
ConFrag introduces a hyperparameter $F$, the number of fragments.
While we simply set $F=4$ for all experiments, we conduct analysis on the effect of using different $F$, as shown in Fig.~\ref{fig:discussion_main}(b).
On SHIFT15M-B dataset, the performance is relatively stable across different fragment numbers.
On IMDB-Clean-B, a small declining trend in performance is observed as the number of fragments increases.
This decrease is likely attributed to a finer division of the training data among feature extractors, ultimately leading to overfitting and reduced generalization capabilities.
Appendix~\ref{subsec:fragment_numbers} provides further analysis of the fragment number.

\begin{table}[t]
    \hspace{-2.5pt}
    \begin{minipage}{.55\textwidth}
        \centering
        \begin{small}
        \setlength{\tabcolsep}{4.2pt}
        \caption{\textbf{Ablation of Mixture of Neighboring Fragments.}
        MRAE on the IMDB-Clean-B dataset (lower is better).
        }
        \label{tab:ablation_mnf}
        \begin{tabular}{cccccc}
            \toprule
            & & & \multicolumn{1}{c}{symmetric}    &\multicolumn{2}{c}{Gaussian} \\
            $\alpha_f^\text{self}$ & $\alpha_f^\text{ngb}$ & $\mathcal{S}$ & 40 & 30 & 50\\
            \midrule
            \checkmark &            & $\mathcal{S}^p\cup \mathcal{S}^r$  & 16.97 & 17.53 & 38.18 \\
                    & \checkmark & $\mathcal{S}^p\cup \mathcal{S}^r$  & 22.22 & 22.77 & 46.36 \\
            \checkmark & \checkmark & $\mathcal{S}^p$                    & 14.18 & 15.84 & 33.07 \\
            \checkmark & \checkmark & $\mathcal{S}^r$                    & 14.06 & 16.94 & 39.85 \\
            \checkmark & \checkmark & $\mathcal{S}^p\cap \mathcal{S}^r$  & 13.08 & 16.18 & 34.23 \\
            \checkmark & \checkmark & $\mathcal{S}^p\cup \mathcal{S}^r$  & 12.64 & 15.70 & 33.36 \\

            \bottomrule
        \end{tabular}
        \end{small}
    \end{minipage}
    \hspace{5pt}
    \begin{minipage}{.45\textwidth}
        \vspace{-3.7pt}
        \caption{\textbf{Parameter size comparison}. regression: parameters for regression, noise: parameters to mitigate noisy labels, ``others": SPR, CDR, D2L, C-Mixup, Sigua, Selfie, BMM, DY-S, Superloss.}
        \label{tab:param_compare}
        \centering
        \begin{footnotesize}
        \setlength{\tabcolsep}{3.0pt}
        \begin{tabular}{lccc}
            \toprule
            & regression & noise & total  \\
            \midrule
            RDI & 23.9M & 47.8M & 47.8M \\
            \specialrule{0.1pt}{1pt}{1pt}
            CNLCU & 47.8M & 47.8M & 47.8M \\
            \specialrule{0.1pt}{1pt}{1pt}
            ``others" & 23.9M & 23.9M & 23.9M \\
            \specialrule{0.1pt}{1pt}{1pt}
            ConFrag & 23.9M & 22.8M & 46.7M \\
            \bottomrule
        \end{tabular}
        \end{footnotesize}
    \end{minipage}
\end{table}

\textbf{Ablation analysis on mixture of neighboring fragments.}
In Table~\ref{tab:ablation_mnf}, we conduct an ablation analysis of the Mixture of neighboring fragments (\S~\ref{subsec:mixture_of_contrasing_fragments}).
When evaluating neighborhood agreement based solely on either the agreement of the current fragment ($\alpha_f^\text{self}$) or the neighboring fragment's agreement ($\alpha_f^\text{ngb}$),
the ablation reveals that relying on the current fragment's agreement alone ($\alpha_f^\text{self}$) exhibited relatively stronger performance.
Nevertheless, this approach still fell short of achieving a satisfactory level compared to considering both agreements, as defined in Eq.~\ref{eq:na_final}.

Next, as we consider sample selection based on two variants of agreements, the \textit{predictive} one utilizing the feature extractor's binary classifier and the \textit{representational} one using $K$-nearest neighbors on the learned feature space
(referred to as the selected sample sets $\mathcal{S}^p$ and $\mathcal{S}^r$ respectively), we conduct an ablation study on these selected sample sets.
This involves evaluating the results when determining the final selected sample set ($\mathcal{S}$) either individually, at the intersection, or at the union of $\mathcal{S}^p$ and $\mathcal{S}^r$.
Overall, in line with ConFrag, the union of sets ($\mathcal{S}^p \cup \mathcal{S}^r$) proves to be the most effective strategy.

\textbf{Parameter size comparison.}
Table~\ref{tab:param_compare} compares the number of parameters of ConFrag and baselines on the ResNet-based age prediction datasets. 
A thorough description of the ConFrag architecture is in Appendix~\ref{subsec:training_details}.
It is worth noting that each of the ConFrag's feature extractors for noise mitigation employs a much fewer number of parameters than the downstream regression task (\eg 48\% in age prediction datasets). 
The total number of parameters of each method varies, as some share parameters for regression as well as noise mitigation while others, such as ConFrag, do not.
Nevertheless, ConFrag uses fewer total parameters than CNLCU-H and RDI.


\section{Conclusion}\label{sec:conclusion}
\vskip -0.05in
To address the problem of noisy labeled regression, we introduce the Contrastive Fragmentation framework (ConFrag). 
The framework partitions the label space and identifies the most contrasting pairs of fragments, thereby training a mixture of feature extractors over contrastive fragment pairs.
This mixture is leveraged for clean selection based on neighborhood agreements.
Extensive experiments on six curated datasets on three domains with different levels of symmetric and Gaussian noise demonstrate that our framework performs superior selection and ultimately leads to a better regression performance than fourteen state-of-the-art models.
Given its foundation in the Mixture of Experts model, the parameter size of ConFrag linearly grows with an increase in the number of fragments. 
We acknowledge this as a potential avenue for future research.

\section*{Acknowledgement}
We express our gratitude to the members of the Vision and Learning Lab at Seoul National University for their valuable feedback on the manuscript.
In particular, we would like to acknowledge Jaekyeom Kim, Soochan Lee, Jinseo Jeong, Wonkwang Lee, and Sehun Lee.
This work was supported by 
LG AI Research,
Institute of Information \& Communications Technology Planning \& Evaluation (IITP) grant funded by the Korea government (MSIT) (No.~RS-2019-II191082, SW StarLab), 
Institute of Information \& communications Technology Planning \& Evaluation (IITP) grant funded by the Korea government (MSIT) (No.~RS-2022-II220156, Fundamental research on continual meta-learning for quality enhancement of casual videos and their 3D metaverse transformation), 
the National Research Foundation of Korea (NRF) grant funded by the Korea government (MSIT) (No.~2023R1A2C2005573), 
and Institute of Information \& communications Technology Planning \& Evaluation (IITP) grant funded by the Korea government (MSIT) (No.~RS-2021-II211343, Artificial Intelligence Graduate School Program (Seoul National University)).  
Gunhee Kim is the corresponding author.
\bibliography{ref}
\bibliographystyle{plainnat}



\newpage
\appendix

\section{Appendix: Table of Contents}

The Appendix enlists the following additional materials. 
\begin{enumerate}[label=\Roman*.]
    \item Limitations. \S~\ref{sec:limitations}
    \item Broader Impacts. \S~\ref{sec:broader_impacts}
    \item Theory of ConFrag. \S~\ref{sec:theory}
        \begin{enumerate}[label=\roman*.]
            \item Classification versus Regression for Feature Learning~\ref{subsec:classification_vs_regression}
            \item Fragmentation and Neighborhood Jittering~\ref{subsec:fragmentation_and_neighborhood_jittering}
            \item Prediction Depth Analysis~\ref{subsec:prediction_depth_analysis}
        \end{enumerate}
    \item Extended Related Work. \S~\ref{subsec:related_work}
        \begin{enumerate}[label=\roman*.]
            \item Continuously Ordered Correlation of Labels and Features~\ref{subsec:continuity_related_works}
            \item Noisy Label in Object Detection~\ref{subsec:object_detection_related_works}
            \item Transition Matrix based Methods~\ref{subsec:transition_matrix_related_works}
            \item Combination with Contrastive Learning~\ref{subsec:contrastive_learning_related_worls}
        \end{enumerate}
    \item Experiment Details. \S~\ref{sec:exp_details}
        \begin{enumerate}[label=\roman*.]
            \item Dataset Curation Details~\ref{subsec:dataset_curation}
            \item Baseline Details~\ref{subsec:baselines}
            \item ConFrag Training Details~\ref{subsec:training_details}
            \item Random Gaussian Noise~\ref{subsec:random_gaussian}
            \item Computation Resource~\ref{subsec:computation_resource}
        \end{enumerate}
    \item Extended Results \& Analyses. \S~\ref{sec:results_analysis}
        \begin{enumerate}[label=\roman*.]
            \item UTKFace Results~\ref{subsec:utkface}
            \item Fragment Number Analysis~\ref{subsec:fragment_numbers}
            \item Hyperparameter Analysis~\ref{subsec:hyperparameter}
            \item Fragment Pairing Analysis~\ref{subsec:contrast_combination}
            \item Closed-Set versus Open-Set Noise~\ref{subsec:disruptive_anomaly_noise}
            \item Analysis of Samples on the Bounday versus Center of Fragment~\ref{subsec:boundary_center}
            \item Ablation \& Combination Analysis~\ref{subsec:ablation}
            \item Discretized Baselines~\ref{subsec:discrete_baselines}
            \item Comparison of Neighborhood Jittering and Other Regularization Methods~\ref{subsec:jitter_comparison}
            \item Extended Selection Rate/ERR/MRAE Comparison and Analysis~\ref{subsec:ERR}
            \item Variance Across Random Seeds~\ref{subsec:variance}
            \item Standard Mean Absolute Error~\ref{subsec:mae}
        \end{enumerate}
    \item ConFrag Pseudo Code (Algorithm~\ref{alg:fragmented_selection})
\end{enumerate}

\section{Limitation}\label{sec:limitations}
A key limitation of ConFrag lies in its foundational reliance on the Mixture of Experts (MoE) model~\citep{jacobs1991MoE}. 
Specifically, integrating MoEs with deep learning introduces notable scalability challenges, both computationally and in memory usage~\citep{zuo21, zoph22, zhang21b}. 
To address the memory concern, ConFrag currently employs more compact feature extractors. Nevertheless, a prominent inefficiency stems from expert redundancy in MoEs' parameters~\citep{zuo21}. 
Some approaches to mitigate this include distilling into sparse MoE models, employing pruning, and subsequently compressing to decrease parameter size~\citep{yjkim21,fedus21}. 
There are also emerging strategies centered on parameter sharing, leveraging matrix product operators (MPO) decomposition~\citep{gao20mpo, gao22mpo} and parameter-efficient fine-tuning~\citep{zadouri23}. 
Of these, we believe the avenue of parameter sharing holds special promise when combined with ConFrag; the inherent positive feature correlation in regression problems amplifies the advantages of this approach.
Also, as in MoEs, ConFrag introduces new hyperparameter, the number of experts (the number of fragments $F$ in ConFrag's case).

In its current form, ConFrag facilitates simultaneous training of both the feature extractors and the subsequent regression task, either on a per-batch or per-epoch basis. 
However, a wealth of research exists that could further optimize ConFrag's scalability. These span from improving training efficiency~\citep{he21fastmoe,zoph22,lepikhin21,lewis21} to enhancing inference capabilities~\citep{zhang21b, fedus21}.

\section{Broader Impacts}\label{sec:broader_impacts}

In the era of deep learning, the need for large datasets increases, yet it is expensive to obtain large dataset with high-quality annotated labels.
An alternative solution is to collect labels using automated labeling methods, such as web crawling.
However, these methods inevitably introduce noisy labels.

This work proposes a method for mitigating the negative effect of such label noise in regression, which can save time and money spent on collecting high-quality labels for many applications, bringing positive impact on science, society, and economy.
However, since the method reduces the need for accurate labeling, it may have potential negative effect on the salaries of label workers.

\section{Theory of ConFrag}\label{sec:theory}
We present several theoretical justifications that enhance the performance of ConFrag.

\subsection{Classification versus Regression for Feature Learning}
\label{subsec:classification_vs_regression}

During the learning process, deep neural networks aim to maximize the mutual information between the learned representation, denoted as \(Z\), and the target variable, denoted as \(Y\). 
The mutual information between these two variables can be defined as \(I(Z; Y) = H(Z) - H(Z | Y)\). A high value of \(I(Z; Y)\) is indicative of a high marginal entropy \(H(Z)\). 
Achieving this dual objective is accomplished in classification~\citep{boudiaf2020}.

However, \citet{zhang2023} have shown that regression primarily focuses on minimizing \(H(Z | Y)\) while disregarding \(H(Z)\). 
This results in a relatively lower marginal entropy for the learned representation \(Z\) and ultimately leads to performance deficits in comparison to classification.

To experimentally show that this theoretical result also applies to ConFrag, we replace classification-based expert feature extractor learning with regression-based one, where each expert feature extractor is trained with regression loss on its respective fragment pair dataset.
We name this variant ConFrag-R.
In ConFrag-R, self-agreement is defined using distances to the mean of each fragment in the contrasting pair $(f, f^+)$:
\begin{align}\label{eq:nar}
\alpha^\text{self}_f &= \left[ |\bar{Y}_f - h(x; \theta_{f_,f^+})| < |\bar{Y}_{f^+} - h(x; \theta_{f_,f^+})| \right],
\end{align}
where $\bar{Y}_f$ is the average of fragment $f$'s labels, and $h_{R}(\cdot)$ is the regression function output.
As in ConFrag, ConFrag-R also utilizes $K$-nearest neighbor-based classification for computing another variant of self-agreement. 
The results in Table~\ref{tab:classification_vs_regression} show that using classification for feature learning outperforms using regression (ConFrag-R) in all datasets.

\begin{table*}[th!]
    \caption{\textbf{Comparison between ConFrag and ConFrag-R: Mean Relative Absolute Error (\%)} to the noise-free trained model on the AFAD-B, IMDB-Clean-B, SHIFT15M-B, and MSD-B dataset.
    Lower is better. 
    The results are the mean of three random seed experiments.
    }
    \begin{center}
    \begin{small}
    \setlength{\tabcolsep}{4.2pt}
    \begin{tabular}{lcccccccccccc}
        \toprule
        &\multicolumn{6}{c}{AFAD-B}         &\multicolumn{6}{c}{IMDB-Clean-B}
        \\\cmidrule(lr){2-7}\cmidrule(lr){8-13}
        &\multicolumn{4}{c}{symmetric}    &\multicolumn{2}{c}{Gaussian} &\multicolumn{4}{c}{symmetric} &\multicolumn{2}{c}{Gaussian}
        \\\cmidrule(lr){2-5}\cmidrule(lr){6-7}\cmidrule(lr){8-11}\cmidrule(lr){12-13} 
        noise rate  & 20 & 40 & 60 & 80 & 30 & 50 & 20 & 40 & 60 & 80 & 30 & 50 \\
        \midrule
        Vanilla & 9.37 & 20.27 & 30.65 & 43.09 & 28.77 & 39.03 & 16.18 & 32.05 & 53.13 & 76.35 & 26.89 & 50.28 \\
        \specialrule{0.1pt}{1pt}{1pt}
        \textbf{ConFrag-R} & 4.97 & 13.93 & 27.85 & 37.19 & 21.93 & 33.90 & 8.74 & 22.73 & 44.29 & 68.14 & 21.74 & 46.93 \\
        \textbf{ConFrag}  & \textbf{2.74} & \textbf{8.16} & \textbf{15.91} & \textbf{34.42} & \textbf{17.49} & \textbf{27.31} & \textbf{5.08} & \textbf{12.64} & \textbf{27.26} & \textbf{61.24} & \textbf{15.70} & \textbf{33.36} \\
        \bottomrule
        \\
        \toprule
        &\multicolumn{6}{c}{SHIFT15M-B}         &\multicolumn{6}{c}{MSD-B}
        \\\cmidrule(lr){2-7}\cmidrule(lr){8-13}
        &\multicolumn{4}{c}{symmetric}    &\multicolumn{2}{c}{Gaussian} &\multicolumn{4}{c}{symmetric} &\multicolumn{2}{c}{Gaussian}
        \\\cmidrule(lr){2-5}\cmidrule(lr){6-7}\cmidrule(lr){8-11}\cmidrule(lr){12-13} 
        noise rate & 20 & 40 & 60 & 80 & 30 & 50 & 20 & 40 & 60 & 80 & 30 & 50 \\
        \midrule
        Vanilla            & 9.11 & 17.96 & 27.02 & 36.34 & 6.54 & 15.16 & 8.23 & 18.43 & 31.67 & 45.85 & 6.96 & 15.74 \\
        \specialrule{0.1pt}{1pt}{1pt}
        \textbf{ConFrag-R} & 4.18 & 9.59 & 16.21 & 25.76 & 4.96 & 10.90 & 0.77 & 5.68 & 13.63 & 30.05 & 2.79 & 6.87 \\
        \textbf{ConFrag} & \textbf{2.46} & \textbf{6.18} & \textbf{10.68} & \textbf{19.04} & \textbf{3.66} & \textbf{8.09} & \textbf{0.57} & \textbf{4.94} & \textbf{11.22} & \textbf{23.41} & \textbf{2.39} & \textbf{6.49} \\
        \bottomrule
    \end{tabular}
    \end{small}
    \end{center}
    \label{tab:classification_vs_regression}
\end{table*}

\subsection{Fragmentation and Neighborhood Jittering}
\label{subsec:fragmentation_and_neighborhood_jittering}
ConFrag operates by partitioning data samples into fragments and leveraging trained feature extractors for sample selection through collective modeling. 
We conceptualize this as a Mixture-of-Experts (MoE) model, wherein individual experts specialize in specific problem subspaces through data partitioning~\citep{yuksel2012Moe,masoudnia2014Moe}. 
MoEs possess theoretically advantageous properties with respect to computational scalability and reduction of output variance~\citep{yuksel2012Moe}, contributing to the enhancements observed in ConFrag. 
It is noteworthy that since each network is trained on a distinct training set, MoE effectively mitigates concurrent failures, thereby preventing error propagation among networks and ultimately improving the generalization performance of ConFrag as well~\citep{sharkey1997}.

Additionally, our Neighborhood Jittering leads to a Partially Overlapping Mixture Model~\citep{heller2007}, theoretically enabling the modeling of significantly richer and more intricate hidden representations by accommodating multi-cluster membership, ultimately enhancing the selection and overall performance of ConFrag.

\subsection{Prediction Depth Analysis}
\label{subsec:prediction_depth_analysis}

Prediction depth \citep{baldock21nips} of an example refers to the earliest layer where the layer-wise $K$-nearest neighbor probes of the layer and all the subsequent layers are the same as the model prediction.
In other words, low prediction depth means that the example is easily distinguishable in early layers.
For example, a prediction depth of zero means that data can be predicted at the input level only based on its distances to other data.
Low prediction depth is positively correlated with better prediction consistency, lower learning difficulty, and larger margin.
Due to these traits, some previous works aim at reducing the prediction depth during training for better generalization performance \citep{zhou2021fortuitous, sarfi2023simulated}.

While prediction depth is initially designed as a measure of example difficulty, the mean prediction depth of the dataset can also be used as a measure of dataset difficulty \citep{baldock21nips}.
Also, since early layers generalize while later layers memorize in deep learning \citep{stephenson2021geometry}, the low mean prediction depth of a dataset means it is more generalizable since fewer examples require memorization.

\begin{table*}[th!]
    \caption{\textbf{Comparison of mean prediction depths} of feature extractor learning tasks for all-frag, contrastive fragment pairing, and alternative fragmentation pairings when $F=4$.}
    \begin{center}
    \begin{tabular}{lccc}
        \toprule
         & & \multicolumn{2}{c}{IMDB-Clean-B} 
        \\ \cmidrule(lr){3-4}
        \multicolumn{2}{l}{Fragment pairing} & No noise & Symmetric 40\% \\
        \midrule
        \multicolumn{2}{l}{All-frag ([1-4])} & 6.6291 & 7.2452 \\
        \specialrule{0.7pt}{1pt}{1pt}
        \multirow{2}{*}{\makecell{Contrastive \\ pairing}} & [1, 3] & 3.7752 & 4.8116 \\
        & [2, 4] & 3.7028 & 4.7869 \\
        \specialrule{0.7pt}{1pt}{1pt}
        \multirow{2}{*}{\makecell{Alternative \\ pairing 1}} & [1, 4] & 3.0822 & 4.3307 \\
        & [2, 3] & 4.8215 & 5.3978 \\
        \specialrule{0.7pt}{1pt}{1pt}
        \multirow{2}{*}{\makecell{Alternative \\ pairing 2}} & [1, 2] & 4.8738 & 5.4741 \\
        & [3, 4] & 4.7304 & 5.1828 \\
        \bottomrule
    \end{tabular}
    \end{center}
    \label{tab:pred_depth_f4}
\end{table*}

\begin{table*}[th!]
    \caption{\textbf{Comparison of mean prediction depths} of feature extractor learning tasks for all-frag, contrastive fragment pairing, and alternative fragmentation pairings when $F=6$.}
    \begin{center}
    \begin{tabular}{lccc}
        \toprule
         & & \multicolumn{2}{c}{IMDB-Clean-B} 
        \\ \cmidrule(lr){3-4}
        \multicolumn{2}{l}{Fragment pairing} & No noise & Symmetric 40\% \\
        \midrule
        \multicolumn{2}{l}{All-frag ([1-6])} & 7.2689 & 7.8102 \\
        \specialrule{0.7pt}{1pt}{1pt}
        \multirow{3}{*}{\makecell{Contrastive \\ pairing}} & [1, 4] & 3.8635 & 4.8706 \\
        & [2, 5] & 3.7668 & 4.6382 \\
        & [3, 6] & 3.6840 & 4.5026 \\
        \specialrule{0.7pt}{1pt}{1pt}
        \multirow{3}{*}{\makecell{Alternative \\ pairing 1}} & [1, 2] & 4.8979 & 5.4790 \\
        & [3, 4] & 5.0770 & 5.2453 \\
        & [5, 6] & 4.8010 & 5.0605 \\
        \specialrule{0.7pt}{1pt}{1pt}
        \multirow{3}{*}{\makecell{Alternative \\ pairing 2}} & [1, 6] & 2.9685 & 4.1188 \\
        & [2, 5] & 3.7668 & 4.6382 \\
        & [3, 4] & 5.0770 & 5.2453 \\
        \bottomrule
    \end{tabular}
    \end{center}
    \label{tab:pred_depth_f6}
\end{table*}

Tab.~\ref{tab:pred_depth_f4}--\ref{tab:pred_depth_f6} show the mean prediction depth of all samples for each feature extractor learning task, with and without noise.
The prediction depth for each task is measured using ResNet-18 trained for 20 epochs, at which the model achieves more than 99\% training accuracy.
Following \citet{baldock21nips}, we use $K=30$ for $K$-nearest neighbor probe and did not use data augmentation during training.
The tables show that the binary classification tasks from contrastive pairing achieve much lower prediction depths than the multi-class classification tasks using all fragments (All-frag).
Also, the mean and maximum prediction depths of contrastive pairing are lower than those of alternative pairings, explaining why contrastive pairing outperforms alternative pairings as shown in \S~\ref{sec:discussion}.
Note that the mean prediction depths of the binary classification tasks correlate with the distance between fragments: the larger the distance, the lower the prediction depth tends to be.

\begin{figure*}[th]
\begin{center}
\centerline{\includegraphics[width=0.8\textwidth]{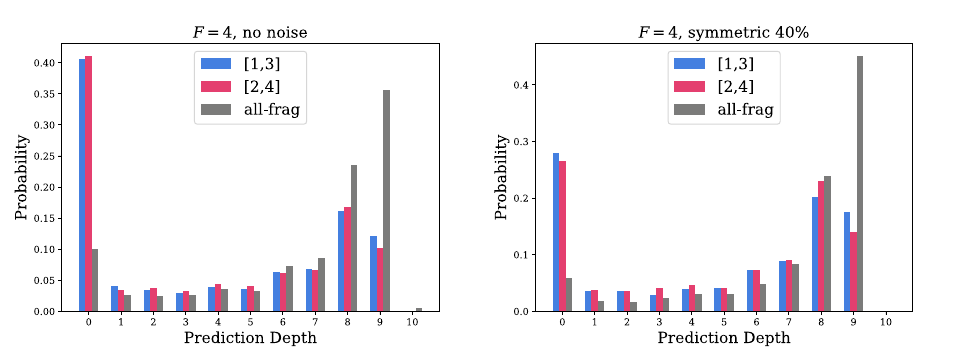}}
\caption{
    \textbf{Probability of prediction depth} for examples in contrastive pairing and all-frag ($F=4$).}
\vskip -0.15in
\label{fig:pred_depth_plot_f4}
\end{center}
\end{figure*}

\begin{figure*}[th]
\begin{center}
\centerline{\includegraphics[width=0.8\textwidth]{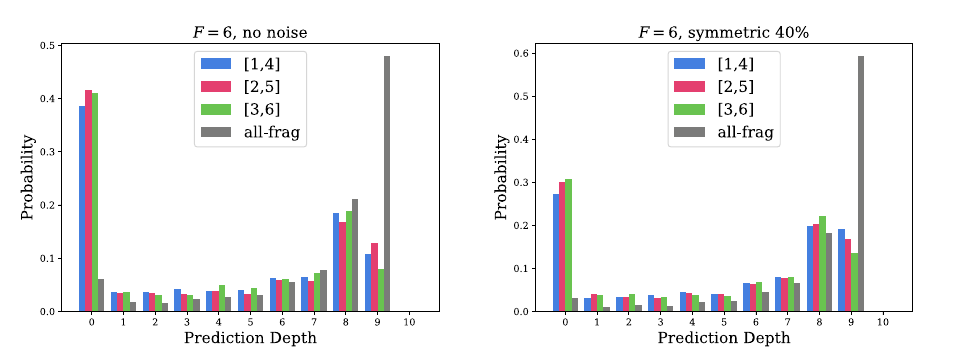}}
\caption{
    \textbf{Probability of prediction depth} for examples in contrastive pairing and all-frag ($F=6$).}
\vskip -0.15in
\label{fig:pred_depth_plot_f6}
\end{center}
\end{figure*}

Fig.~\ref{fig:pred_depth_plot_f4}--~\ref{fig:pred_depth_plot_f6} show the distribution of prediction depth for each task when $F=4$ and $F=6$.
Without noise, about 40\% of data in each contrastive fragment pair has a prediction depth of zero, indicating that two fragments are already much separated at input level.
Even with 40\% symmetric noise, more than 25\% of data in each contrastive fragment pair has a prediction depth of zero.
Meanwhile, the most frequent prediction depth when using all fragments is nine, which means that most data can only be predicted at the last hidden feature level.

While the prediction depths of alternative pairings are lower than those of using all fragments, we observe that they tend to perform worse as shown in Appendix~\ref{subsec:contrast_combination}.
We suspect that this is due to the limitation of mixture models that the individual expert feature extractor may not fully benefit from the full dataset as they model their own disjoint subset.

\section{Extended Related Work}\label{subsec:related_work}

\subsection{Continuously Ordered Correlation of Labels and Features}\label{subsec:continuity_related_works}
One distinctive characteristic of regression problems is their continuous label space, implying a high likelihood of correlation between regions within the feature and label spaces~\citep{yang21ldsfds, gong22rank, zha22supcr}.

Recent research has extensively explored these characteristics, encompassing issues such as label imbalance~\citep{yang21ldsfds, gong22rank}, age estimation~\citep{li19bridge}, contrastive learning~\citep{zha22supcr}, and mixup regularization~\citep{yao22cmixup}. 

\citet{yang21ldsfds} propose label and feature distribution smoothing based on their similarity, while \citet{gong22rank} introduce a regularization term aimed at aligning the rankings of feature-space and label-space neighbors. 
\citet{zha22supcr} employ supervised contrastive learning with a pairing technique based on label distances in mini-batches. 
To adapt MixUp~\citep{zhang18mixup} for regression tasks, \citet{yao22cmixup} recommend interpolating proximal samples within the label space with a higher probability.

Ordinal regression, also known as ranking learning, pertains to predicting ordinal labels based on input data. 
It is noteworthy that ordinal regression methods are adaptable for regression tasks due to the inherent numerical ordering within scalar label spaces. 
Past studies in ordinal regression have successfully addressed various regression challenges, including facial age estimation~\citep{niu16afad, shin2022moving}, monocular depth estimation~\citep{fu2018deep}, and credit rating~\citep{hirk2019multivariate}. 
Some of these methods share common characteristics with our approach, as they discretize continuous labels, effectively converting regression tasks into classification problems~\citep{niu16afad, fu2018deep, shah2022label}. 
Within the framework of ordinal regression, \citet{garg2020robust} propose a loss correction method by estimating the noise transition matrix.

It is important to note that among the previously mentioned methods, only \citet{yao22cmixup} and \citet{garg2020robust} can effectively address noisy label regression problems without the need for additional techniques. 
Additionally, \citet{wang22spr} enhance the scalability of their approach by grouping dissimilar classes within the feature space. 
Our work considers the continuity of labels and features and their correlation in fragmenting and grouping data. 
This approach allows each component to learn distinguishable features and improve sample selection capabilities.

\subsection{Noisy Label in Object Detection}\label{subsec:object_detection_related_works}
Due to the abundance of research on object detection tasks, with bounding box localization being a prominent example of regression tasks,
we have explored the issue of noisy regression within the context of object detection.
In particular, obtaining accurate annotations for object detection is a resource-intensive task, often constrained by limited time, a small number of annotators, or reliance on machine-generated annotations.
These constraints frequently result in label noise, represented as incorrect class assignments or inaccurate bounding box locations.

Various strategies have been developed to address the issue of noisy labels in object detection.
To correct inaccurate bounding box locations, \citet{li2020towards} leverage the discrepancy between two classification heads by emphasizing the objectness of the region.
\citet{liu2022robust} generates object bags using the classifier as guidance, \citet{mao2021noisy} employs center-matching correction, and \citet{schubert2023identifying} drop instances with high region proposal loss on an instance-wise basis.
In scenarios where image-level annotations are available, \citet{gao2018notercnn} employs ensemble learning with two classification heads and a distillation head,
while \citet{shen2020noise} decomposes the problem into foreground and background noise, employing residual learning and bagging-mixup learning.

We also explored the possibility of applying object detection techniques to noisy labeled regression.
However, our analysis revealed that these methods are not well-suited for the broader regression task.
Specifically, \citet{liu2022robust, schubert2023identifying, mao2021noisy} utilize region proposal networks to generate bounding box proposals.
They leverage these proposals to selectively choose clean labels or re-weight the training samples.
However, because this approach necessitates an auxiliary model in the proposal generation process, it cannot be directly applied in the context of regression tasks.

Additionally, \citet{li2020towards, liu2022robust, schubert2023identifying, gao2018notercnn} employ the object detector's classifier to update or assess the quality of bounding boxes.
By evaluating the confidence or consistency of the bounding box through the classification output, this approach helps mitigate the impact of noisy labels.
However, implementing a similar approach in the context of regression tasks would require the inclusion of an auxiliary co-trained task.

\subsection{Transition Matrix based Methods}\label{subsec:transition_matrix_related_works}
Methods based on transition matrices constitute one of the primary approaches for addressing the issue of noisy labels.

Driven by the observation that the clean class posterior, denoted as $p(y^\text{gt}|x)$,
can be inferred from the transition probability and the noisy class posterior, $p(y|x)=T(y|y^\text{gt})p(y^\text{gt}|x)$,
the modification of the loss function enables the construction of a risk-consistent estimator using the estimated transition matrix~\citep{yao2020dual}.

There are many approaches aiming to enhance the estimation of the transition matrix.
These include factorizing it into the product of two matrices by introducing an intermediate class~\citep{yao2020dual}, training the Bayes label transition network~\citep{yang2021estimating},
learning the transition matrix within a meta-learning framework~\citep{wang2020training}, down-weighting less informative features based on $f$-mutual information~\citep{zhu2022beyond},
and adopting a two-head architecture. The latter involves a noisy classifier for simultaneous transition matrix estimation and a clean classifier for statistically consistent training~\citep{kye2021learning}.

Moreover, \citet{xia2020part} explores the utilization of part-dependent transition matrices, combining them to approximate the instance-dependent transition matrix.

In an extended context, \citet{li2022estimating} broadens the problem to include noisy multi-label learning and suggests considering label correlations.

\subsection{Combination with Contrastive Learning}\label{subsec:contrastive_learning_related_worls}
Incorporating unsupervised learning methods proves effective in alleviating label noise, prompting the integration of noisy label mitigation techniques with unsupervised learning, particularly contrastive learning.

\citet{zhang2021codim} show that the combination of contrastive loss and semi-supervised loss yields successful mitigation of the noisy label problem.

Beyond the application of contrastive learning, other approaches involve selecting confidence pairs and confidence samples~\citep{li2022selective},
leveraging clean probability estimation derived from the relationship between representation clusters and labels~\citep{huang2023twin},
employing class prototypes for weakly-supervised loss~\citep{li2021learning}, and implementing soft-labeling based on the relation between representations and labels~\citep{ortego2020multi}.

Additionally, an approach introduces a contrastive regularization function aimed at preventing adverse effects stemming from noisy labels~\citep{yi2022onleraning}.

\begin{table}[t]
    \caption{\textbf{Dataset Statistics} on the six newly curated balanced datasets for regression: AFAD-B~\citep{niu16afad}, IMDB-Clean-B~\citep{lin2021imdbclean}, IMDB-WIKI-B~\citep{rothe18imdb}, UTKFace-B~\citep{zhifei2017utkface}, SHIFT15M-B~\citep{kimura21shift15m}, MSD-B~\citep{bertin11msd}.}
    \label{tab:dataset_statistics}
    \centering
    \begin{footnotesize}
    \setlength{\tabcolsep}{5pt}
    \begin{tabular}{lccccc}
        \toprule
        Dataset  & range & train & valid & test & total \\
        \midrule
        AFAD-B & [15, 40] & 27647 & 1627 & 3252 & 32526 \\
        \specialrule{0.1pt}{1pt}{1pt}
        IMDB-Clean-B & [15, 66] & 44200 & 2600 & 5200 & 52000 \\
        \specialrule{0.1pt}{1pt}{1pt}
        IMDB-WIKI-B & [15, 65] & 42500 & 2500 & 5000 & 50000 \\
        \specialrule{0.1pt}{1pt}{1pt}
        UTKFace-B & [1, 70] & 10467 & 386 & 787 & 11640 \\
        \specialrule{0.4pt}{1pt}{1pt}
        SHIFT15M-B  & [0, 40000] & 273417 & 16080 & 32180 & 321677 \\
        \specialrule{0.4pt}{1pt}{1pt}
        MSD-B & [1956, 2010] & 25218 & 1512 & 2970 & 29700 \\
        \bottomrule
    \end{tabular}
    \end{footnotesize}
\end{table}

\section{Experiment Details}\label{sec:exp_details}

\subsection{Dataset Curation Details}\label{subsec:dataset_curation}
Table~\ref{tab:dataset_statistics} provides a comprehensive overview of the statistics for the six benchmark datasets meticulously curated for the task of noisy label regression. 
Detailed descriptions of the dataset tailoring process are presented below for clarity.

\textbf{Age prediction datasets (IMDB-Clean-B, AFAD-B, IMDB-WIKI-B, UTKFace-B)}: These datasets are harmonized by achieving a balance across distinct age values. 
This equilibrium is established using a bin sample count threshold (clip value) of 1000 for IMDB-Clean-B and IMDB-WIKI-B, 1251 for AFAD-B, and 200 for UTKFace-B. 
Image inputs are resized to dimensions of $(128\times128)$. 
For the regression task, we consistently employ a ResNet-50 backbone across all models.

\textbf{SHIFT15M-B}: Achieving data balance in this dataset involves a two-step process. 
First, the label space is binned based on a price threshold of \textyen2000. 
Subsequently, data points exceeding the maximum price of \textyen40000 are clipped to remove outliers. 
The binning threshold is set at 16084 sample counts to further ensure balanced representation. 
To standardize the label currency, it is pegged to the U.S. dollar, referencing exchange rates from 2010 to 2020, which coincides with the period when the original clothing item data is collected. 
Notably, this dataset is provided as the penultimate feature of the ImageNet pre-trained VGG-16 model. 
Consequently, we opt for a three-layer MLP architecture with a hidden layer size of [2048, 1024, 512], aligning with recommendations from \citet{papadopoulos22fashion} and \citet{kimura21shift15m}.

\textbf{MSD-B}: Achieving balance in the Million Song Dataset involves setting a threshold of 550 samples per year. 
For all regression models in this context, we adopt a regression backbone rooted in the tabular ResNet structure proposed by \citet{gorishniy21nips}, featuring a hidden dimension of 467.

\textbf{Licenses of existing datasets. }
IMDB-Clean dataset \citep{lin2021imdbclean} is under MIT license.\footnote{\url{https://github.com/yiminglin-ai/imdb-clean}}
SHIFT15M dataset \citep{kimura21shift15m} is under CC BY-NC 4.0 and MIT license.\footnote{\url{https://github.com/st-tech/zozo-shift15m}}
MSD \citep{bertin11msd} song year prediction dataset is under CC BY 4.0 license.\footnote{\url{https://archive.ics.uci.edu/dataset/203/yearpredictionmsd}}
UTKFace dataset \citep{zhifei2017utkface} is available for non-commercial research purposes only.\footnote{\url{https://susanqq.github.io/UTKFace/}}
IMDB-WIKI dataset \citep{rothe2015imdb-wiki,rothe18imdb} is available for academic research purpose only.\footnote{\url{https://data.vision.ee.ethz.ch/cvl/rrothe/imdb-wiki/}}
Unfortunately, the license of AFAD \citep{niu16afad} dataset could not be found.\footnote{\url{https://github.com/John-niu-07/tarball}}

\subsection{Baselines Details}\label{subsec:baselines}
While numerous branches of noisy labeled learning have been explored for classification tasks, our focus in this study centers on the challenging domain of noisy label regression. 
To comprehensively investigate this task, we have conducted an extensive review of the various branches and have selected a set of fourteen baselines that are adaptable to regression. 
It is worth noting that C-Mixup~\citep{yao22cmixup} was originally proposed as a regression baseline.
In the following section, we provide an overview of these selected baselines, offering a broad coverage of diverse approaches to address the noisy label regression problem. 
Additionally, we present detailed descriptions of the experimental settings for each baseline.

\begin{enumerate}
\item D2L~\citep{ma18d2l} for intrinsic dimension exploration. Following the paper, we set $k=20$ and $m=10$ for Local Intrinsic Dimensionality (LID) estimation
and set the LID estimation window as five following the official implementation.
\item CDR~\citep{xia21cdr} for model weight parameter selection, and RDI~\citep{hu20rdiaux} for regularizing the paramter distance from the initialization.
At RDI, we use search space $\lambda \in [0.25, 0.5, 1, 2, 4, 8]$.
\item C-Mixup~\citep{yao22cmixup} for regularization via continuous mixup.
C-Mixup-batch is used in all experiments because of the excessive memory requirement for pairwise distance matrix $P$.
We set the beta distribution variable $\alpha$ as 1.5.
The bandwidth variable $\sigma$ is searched over [0.01, 0.1, 1], following~\citet{yao22cmixup}.
\item SELFIE~\citep{song19b} and AUX~\citep{hu20rdiaux} for refurbishing.
To apply SELFIE to the continuous label, we redefine the concept of uncertainty $F(x;q)$ and
refurbished labels $y^{refurb}$ with the mean and standard deviation.
\begin{align}\label{eq:selfie_uncertainty}
    F(x;q) = \frac{\sigma(H_x(q))}{(\max{(Y)}-\min{(Y)})} < \epsilon
\end{align}
\begin{align}\label{eq:selfie_refurbished_label}
    y^{refurb} = \mu(H_x(q))
\end{align}
where $H_x(q)$ is the prediction history of $x$ from before $q$ epochs, $\epsilon$ is the uncertainty threshold.

For SELFIE, we train 1/4 of the total training epochs for the warm-up phase, following~\citet{song19b}.
The variable $q$ is searched over half of the warm-up epochs and around.
The variable $\epsilon$ is searched over [0.05, 0.10, 0.15, 0.20], following~\citet{song19b}.

For AUX~\citep{hu20rdiaux}, we regularize the auxiliary variable by weight decay 0.0005, reducing the weight by 0.1 at 1/2 and 3/4 of the total training epochs.
The learning rate of the auxiliary variable is set to 0.1 and 0.01.
The variable $\lambda$ is searched over [0.25, 0.5, 1, 2, 4, 8].
\item SPR~\citep{wang22spr} performs penalized regression for selection. It requires some adaptation to regression by ignoring the $\ell_q$
penalty as there is no longer a linearity gap between the scalar output and the final fully connected layer that requires reducing.
Also, we use our fragmentation splits $\{4, 8\}$ to bin the regression data for SPR's parallel optimization.
\item Sigua~\citep{han20sigua} and CNLCU-S/H~\citep{xia22} for small loss selection.
For Sigua, we use $\delta(t)\in[0.3, 0.4]$ and $\gamma=0.01$ and set $T_k$ as 5\% of the total training epochs.
For CNLCU-S/H, we search $\sigma$ and $\tau_{\min}$ in [0.01, 0.1, 1, 10] and set $T_k$ as 5\%.

\item BMM~\citep{arazo19} for selection based on beta mixture model fitting on the loss distribution. 
BMM does hard sampling and trains using the selected samples.
DY-S is a dynamic soft loss. We implemented two versions; the first uses a convex combination as in \citet{reed15} $((1-w)\tilde{y}^c - w\hat{y})^2$. Second, instead of bootstrapping, we dynamically weight the loss using the BMM probability to create a cost-sensitive loss, $(1-w)\ell$.
The $w$ is the mixture clean probability, $\hat{y}$ is the model prediction, $\tilde{y}^c$ is the assigned noisy label, and $\ell$ is the loss.

\item SuperLoss~\citep{castells20} regularization parameter $\lambda$ is searched over [0.01, 0.1, 1, 10] while $\tau$ uses an exponential running average with a fixed smoothing parameter $\alpha = 0.9$.

\item \text{[Incompatible]} CRUST~\citep{mirzasoleiman20crust} for clean coreset selection.
It aims to select a coreset based on \textit{class-wisely gradient clustering}.
For regression, we initially viewed \textit{all data as a single class} and proceeded with coreset selection, but the results were unsatisfactory.
Therefore, we report results based only on the discretized version, demonstrating comparable performances.
We select 1/2 of the total dataset as a coreset.
The distance threshold in calculating clusters is searched over [1, 2, 4].

\item \text{[Incompatible]} OrdRegr~\citep{garg2020robust} for loss correction.
Since no official implementation is provided, we implemented it with cross-entropy loss for ordinal regression.
Importantly, we failed to find accurate noise rate estimation using their suggested methods.
Even when considering the transition matrix with the actual noise rate, the loss correction algorithm proved ineffective in our benchmark tests.

\end{enumerate}

\subsection{ConFrag Training Details}\label{subsec:training_details}

ConFrag employs the Cosine Annealing Learning rate~\citep{loshchilov17iclr} with a minimum learning rate of $\eta_{min}=0$.
The optimization is carried out using the Adam optimizer~\citep{kingma15adam}.
For the $K$-nearest neighbors-based prediction, we experiment with various values of $K$, specifically choosing from the set $[3, 5, 7]$. 
The number of fragments, denoted as $F$, remains constant at four throughout all the experiments. 
To determine the buffer range for jittering, we conduct a search over values within the range $[0, 0.05, 0.1]$.

Some dataset-specific hyperparameters exist:
\begin{itemize}
    \item Age prediction task datasets, IMDB-Clean-B~\citep{lin2021imdbclean}, AFAD-B~\citep{niu16afad}, IMDB-WIKI-B~\citep{rothe18imdb}, and UTKFace-B~\citep{zhifei2017utkface}, train for 120 epochs
    with a learning rate of 0.001. 
    Each feature extractor employs the ResNet-18 architecture, which contains only 48\% of the parameters found in ResNet-50, the architecture utilized for the regressor.
    \item Clothing price estimation task dataset SHIFT15M-B~\citep{kimura21shift15m} trains for 40 epochs
    with a learning rate of 0.0001.
    MLP with hidden dimensions [1024, 512, 256] is deployed for feature extractors, and the parameter size is 44\% of the regressor.
    \item Music year production task dataset MSD-B~\citep{bertin11msd} trains for 20 epochs
    with a learning rate of 0.0001. 
    Similar to the regression backbone, the feature extractor model is the tabular ResNet structure\citep{gorishniy21nips}, and the hidden dimension is reduced to 256.
\end{itemize}

\begin{figure}[t]
\begin{center}
\centerline{\includegraphics[width=0.8\columnwidth]{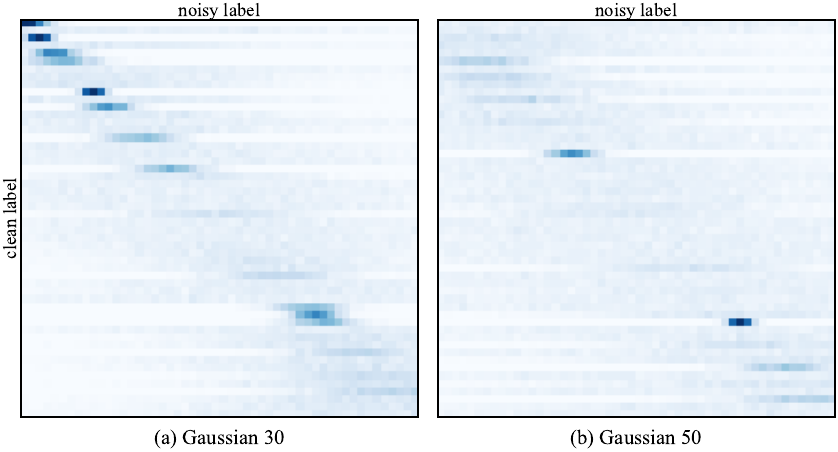}}
\caption{\textbf{Random Gaussian Noise.} (a) Gaussian noise injected from the uniformly sampled random standard deviation between $[1,30]$. 
(b) Gaussian noise injected from uniformly sampled random standard deviation between $[1,50]$.}
\label{fig:gaussian_depict}
\end{center}
\end{figure}

\subsection{Random Gaussian Noise}\label{subsec:random_gaussian}
Fig.~\ref{fig:gaussian_depict} illustrates the application of random Gaussian noise within the label space of IMDB-Clean-B~\citep{lin2021imdbclean}. 
The procedure for injecting noise is akin to the approach employed by \citet{yao22cmixup}, where Gaussian noise is applied to every unique label within the training samples. 
Specifically, \citet{yao22cmixup} sets the standard deviation of the Gaussian noise as a fixed 30\% of the range of the label space corresponding to the dataset. 
In contrast, our noise injection method introduces an element of stochasticity, allowing for variable levels of deviation for each unique label.

To achieve this variability, we employ uniform sampling from the minimum and maximum values specific to each label's domain. 
For instance, in the context of an age prediction task, we assume minimum and maximum values of 0 and 100, respectively. 
However, in cases where the label domain lacks clarity (\eg for a variable like `price'), we utilize the minimum and maximum label values provided by the dataset itself.

It is important to highlight that baselines with known noise rates, such as CNLCU-S/H, Sigua, and Selfie, are incapable of dealing with Gaussian noise.
Given that these baselines employ a heuristic approach to control selection rates through $(1 - \text{noise rate})$,
they prove ineffective when exposed to Gaussian noise, as it introduces noise to all samples, thereby resulting in a nearly 100\% noise rate.
Hence, we create a \emph{soft noise rate} to be used by them for selection.
This is done by calculating an updated noise rate, assuming that the Gaussian noise injected samples that fall within an acceptable variance of the original ground-truth label are clean
(the acceptable variance is set to equal the label length/size of a single fragment).

\subsection{Computation Resource}\label{subsec:computation_resource}

For implementation, we use Python 3.9 and PyTorch 1.13.1.
All experiments are conducted using NVIDIA Quadro 6000 24GB RAM GPUs.
The required computation time for experiments differs depending on the dataset.
Below, we report the computation time for each dataset.

\textbf{AFAD-B. }
On the AFAD-B dataset, the average computation time of ConFrag and Co-ConFrag are approximately 2.5 GPU hours and 5 GPU hours, respectively.
The computation time of baselines ranges from 1.25 GPU hours to 4 GPU hours.
About 650 GPU hours were required to produce the AFAD-B part of Tab.~\ref{tab:main_mrae}.

\textbf{IMDB-Clean-B. }
On the IMDB-Clean-B dataset, the average computation time of ConFrag and Co-ConFrag are approximately 3.5 GPU hours and 5.5 GPU hours, respectively.
The computation time of baselines ranges from 2 GPU hours to 5 GPU hours.
About 970 GPU hours were required to produce the IMDB-Clean-B part of Tab.~\ref{tab:main_mrae}.

\textbf{SHIFT15M-B. }
On the SHIFT15M-B dataset, the average computation time of ConFrag and Co-ConFrag are approximately 1 GPU hour and 1.2 GPU hours, respectively.
The computation time of baselines ranges from 6 GPU minutes to 1 GPU hour.
About 100 GPU hours were required to produce the SHIFT15M-B part of Tab.~\ref{tab:main_mrae}.

\textbf{MSD-B. }
On the MSD-B dataset, the average computation time of ConFrag and Co-ConFrag are approximately 4 GPU minutes and 5 GPU minutes, respectively.
The computation time of baselines ranges from 1 GPU minute to 4 GPU minutes.
About 9 GPU hours were required to produce the MSD-B part of Tab.~\ref{tab:main_mrae}.

Additionally, further computation was required for analysis and experiments in \S~\ref{sec:discussion} and Appendix~\ref{sec:results_analysis}.

\section{Extended Results \& Analysis}\label{sec:results_analysis}
We conduct supplementary experiments and analyses of UTKFace dataset, parameter sizes, fragment numbers ($F$), other hyperparameters ($K$, $J$), fragment pairing, and the impact of closed-set and open-set noise.
Furthermore, we present ablation analyses, comparisons with discretized baselines, baseline performance evaluations considering Selection rate and ERR, variance assessments,
and the obtained MAE results.

\subsection{UTKFace Results}\label{subsec:utkface} 
Table~\ref{tab:utkface} presents a comparison of MRAE values between various baseline methods and our proposed approach on the balanced UTKFace dataset~\citep{zhifei2017utkface}, UTKFace-B. 
We experiment under four different symmetric noise conditions of symmteric 20\%, 40\%, 60\% and 80\% noise rates.
Both ConFrag and Co-ConFrag demonstrate superior performance across all experiments when compared to the fourteen baseline methods.

\begin{table*}[t]
    \centering
    \caption{Comparison of MRAE (\%) on UTKFace-B datasets with symmetric noise.}
    \begin{small}
    \setlength{\tabcolsep}{4.2pt}
    \begin{tabular}{lcccc}
    \toprule
    &\multicolumn{4}{c}{\textbf{UTKFace-B}} \\
    \cmidrule(lr){2-5}
    &\multicolumn{4}{c}{symmetric} \\
    \cmidrule(lr){2-5}
    \textbf{} & 20 & 40 & 60 & 80 \\
    \midrule
    Vanilla &  37.59 & 53.96 & 82.88 & 115.49 \\
    AUX &  25.53 & 48.01 & 84.78 & 118.31 \\
    BMM &  49.21 & 76.25 & 88.87 & 139.01 \\
    CDR & 32.87 & 50.56 & 83.41 & 121.36 \\
    C-Mixup &  17.76 & 34.00 & 74.29 & 117.68 \\
    CNLCU-H &  \textcolor{blue}{5.75} & 20.75 & 43.44 & 121.55 \\
    CNLCU-S &  20.29 & 36.69 & 43.44 & 121.55 \\
    D2L & 38.28 & 51.22 & 99.19 & 122.50 \\
    DY-S & 22.73 & 31.07 & 58.05 & 113.21 \\
    RDI & 43.49 & 49.44 & 75.67 & 122.40 \\
    Selfie & 30.47 & 44.62 & 99.67 & 130.97 \\
    Sigua &  10.86 & 19.58 & 52.37 & 128.06 \\
    SPR &  28.05 & 48.07 & 85.18 & 120.58 \\
    Superloss & 9.12 & 22.10 & 55.78 & 115.78 \\
    \specialrule{0.7pt}{1pt}{1pt}
    ConFrag &  6.28 & \textcolor{blue}{14.30} & \textcolor{blue}{34.09} & \textcolor{blue}{83.03} \\
    Co-ConFrag & \textcolor{red}{-2.22} & \textcolor{red}{8.88} & \textcolor{red}{25.46} & \textcolor{red}{74.78} \\
    \bottomrule
    \end{tabular}
    \end{small}
    \label{tab:utkface}
\end{table*}


\begin{figure*}[th]
\begin{center}
\centerline{\includegraphics[width=\textwidth]{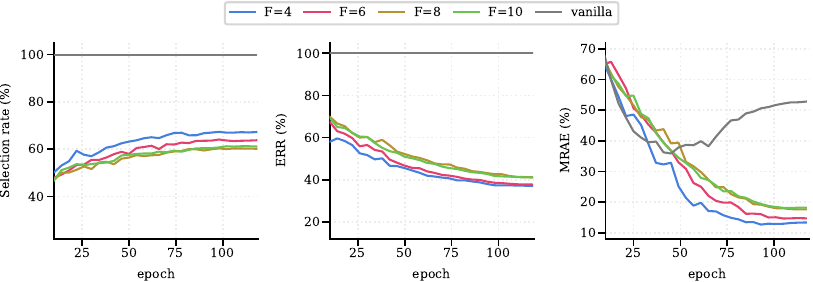}}
\caption{\textbf{Fragment number analysis} compares the Selection rate, ERR and MRAE on IMDB-Clean-B with symmetric 40\% noise.}\label{fig:split_number}

\end{center}
\end{figure*}

\begin{figure*}[th]
\begin{center}
\centerline{\includegraphics[width=\textwidth]{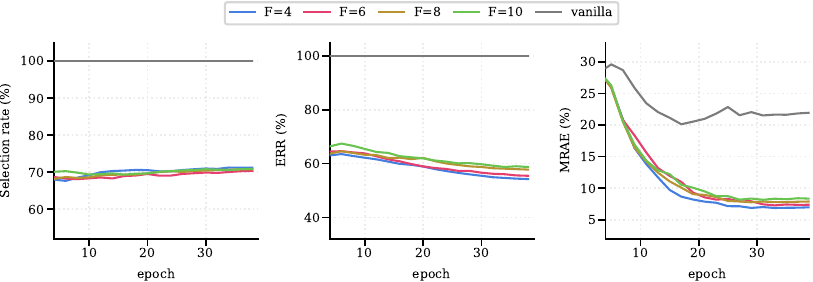}}
\caption{\textbf{Fragment number analysis} compares the Selection rate, ERR and MRAE on SHIFT15M-B with symmetric 40\% noise.}\label{fig:split_number_shift}

\end{center}
\end{figure*}


\subsection{Fragment Number Analysis}\label{subsec:fragment_numbers} 

In Fig.~\ref{fig:split_number} and \ref{fig:split_number_shift}, we undertake an examination of various fragment numbers within the context of symmetric 40\% noise, using the IMDB-Clean-B and SHIFT15M-B datasets as benchmarks. 
Our evaluation criteria encompass the Selection rate, Error Residual Rate (ERR), and Mean Relative Absolute Error (MRAE).
The number of fragments is chosen from $F\in[4, 6, 8, 10]$.
To address scenarios with a smaller fragment number, we examine cases where $F=1$ or $2$.
When $F=2$, a fragment $f$ that satisfies self-agreement (Eq.~\ref{eq:self_agreement}) does not meet the criteria for neighbor-agreement ($\alpha^\text{ngb}_f$ in Eq.~\ref{eq:na_final}),
as the agreement relies on comparing the distribution of fragment $f$ and its contrasting pair $f^+$.
Consequently, the unified neighborhood agreement (Eq.~\ref{eq:na_final}) consistently yields a value of $0$.
On the other hand, defining a contrasting pair is not feasible when $F=1$.
Instead, we present a plot of the vanilla baseline to illustrate the case when $F=1$ without utilizing ConFrag.

The results reveal that the MRAE of the vanilla model initially decreases during the early epochs as it learns patterns from clean samples.
However, as the model starts to memorize noisy samples, the MRAE degrades.
In contrast, ConFrag consistently mitigates the impact of noisy samples across all plots ($F\in[4,6,8,10]$) when compared to the vanilla baseline.

We also observe a declining trend in performance as the number of fragments increases in the case of IMDB-Clean-B. 
In contrast, SHIFT15M-B exhibits relatively stable performance across different fragment numbers. 
This decrease in performance with an increased number of fragments is likely attributed to a finer division of the training data among feature extractors, ultimately leading to overfitting and reduced generalization capabilities.

\begin{figure*}[th]
\begin{center}
\centerline{\includegraphics[width=\textwidth]{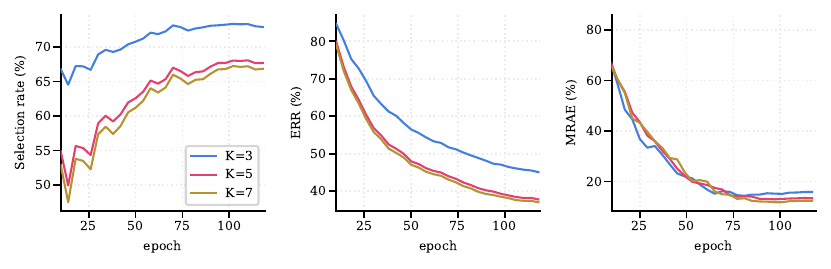}}
\vskip -0.15in
\caption{
    \textbf{Hyperparameter K analysis} compares the Selection rate, ERR and MRAE on IMDB-Clean-B with symmetric 40\% noise.}
    \label{fig:analysis_imdb_k}
\vskip -0.15in
\end{center}
\end{figure*}

\begin{figure*}[th]
\begin{center}
\centerline{\includegraphics[width=\textwidth]{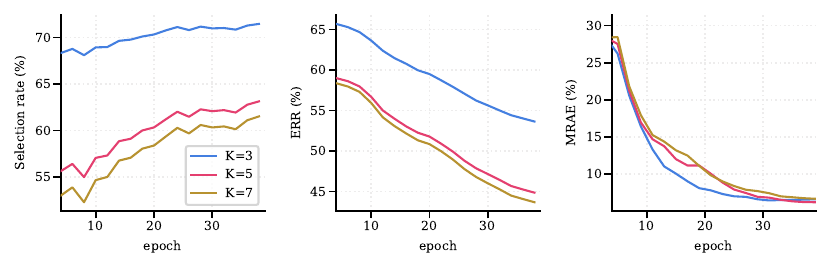}}
\vskip -0.15in
\caption{
    \textbf{Hyperparameter K analysis} compares the Selection rate, ERR and MRAE on SHIFT15M-B with symmetric 40\% noise.}
\vskip -0.15in
\label{fig:analysis_shift15m_k}
\end{center}
\end{figure*}

\begin{figure*}[th]
\begin{center}
\centerline{\includegraphics[width=\textwidth]{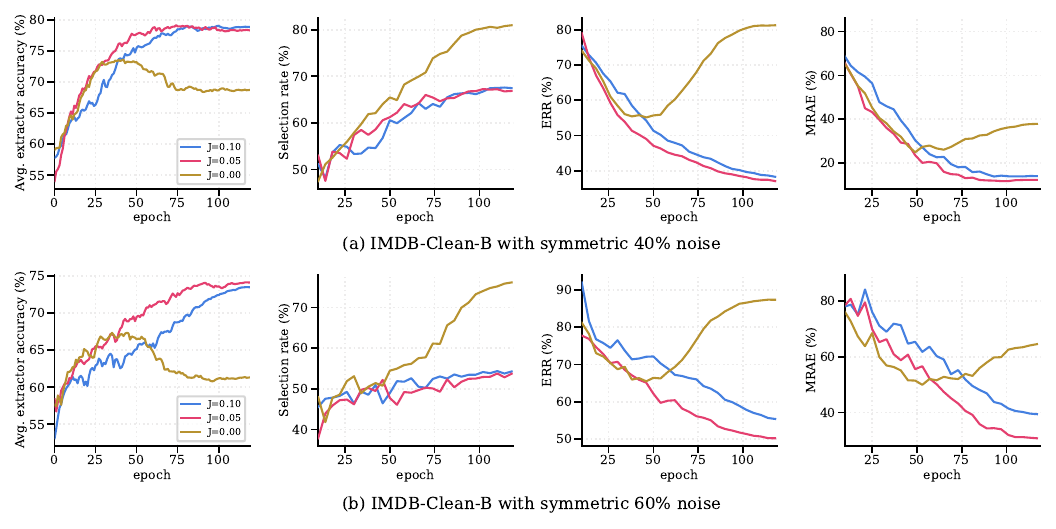}}
\vskip -0.15in
\caption{
    \textbf{Hyperparameter J analysis} compares the average accuracy of feature extractors, the Selection rate, ERR and MRAE on IMDB-Clean-B with symmetric 40\%, 60\% noise.}
\label{fig:analysis_imdb_j}
\end{center}
\end{figure*}

\begin{figure*}[th]
\begin{center}
\centerline{\includegraphics[width=\textwidth]{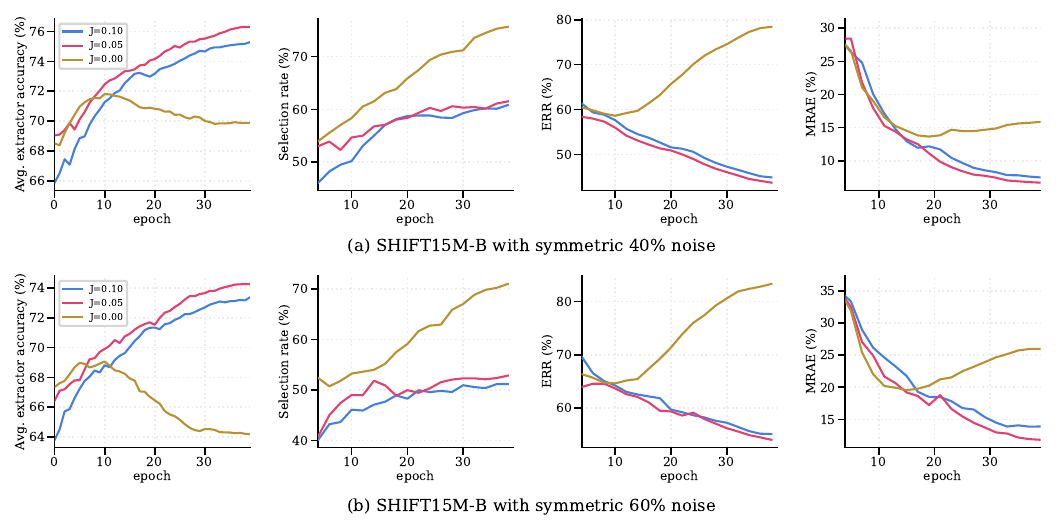}}
\vskip -0.2in
\caption{
    \textbf{Hyperparameter J analysis} compares the average accuracy of feature extractors, the Selection rate, ERR and MRAE on SHIFT15M-B with symmetric 40\%, 60\% noise.}
\label{fig:analysis_shift15m_j}
\end{center}
\end{figure*}

\subsection{Hyperparameter Analysis}\label{subsec:hyperparameter}
The hyperparameter $K$ is used for $K$-nearest neighbor classification when assessing self/neighbor agreement from a representational perspective.
As shown in Fig.~\ref{fig:analysis_imdb_k}, \ref{fig:analysis_shift15m_k}, with an increase in the value of $K$, the criteria for agreement become more stringent.
Consequently, as the value of $K$ increases, a greater number of confident samples are selected, resulting in a reduction in the Selection rate and ERR. 

The hyperparameter $J$ controls the buffer range for jittering, which, in turn, determines the level of regularization applied via neighborhood jittering.
Increasing the value of $J$ results in stronger regularization, effectively preventing overfitting. 
However, excessive regularization, as observed when $J = 0.10$, may result in adverse effects during training.
Specifically, in Fig.~\ref{fig:analysis_imdb_j}(a), the feature extractors exhibit similar convergence patterns when $J = 0.05$ or $J = 0.10$. 
Consequently, comparable performance is observed in Selection Rate and MRAE.
Yet, in Fig.~\ref{fig:analysis_imdb_j}(b), the ERR of $J = 0.05$ is smaller than that of $J = 0.10$, leading to improved MRAE performance for $J=0.05$. 
Similar effects are observed in the SHIFT15M dataset, as depicted in Fig.~\ref{fig:analysis_shift15m_j} (SHIFT15M-B).

\begin{figure*}[th]
\begin{center}
\centerline{\includegraphics[width=\textwidth]{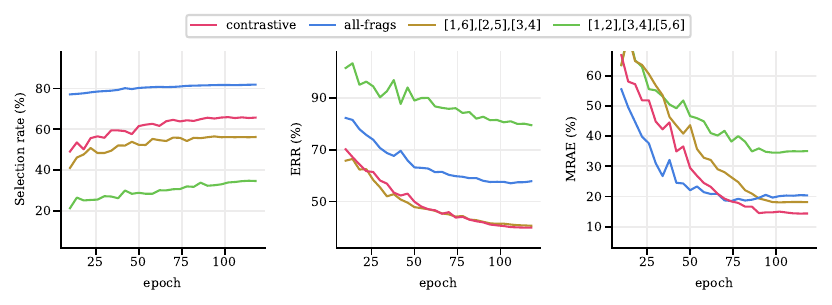}}
\vskip -0.2in
\caption{
\textbf{Fragment pairing analysis} compares contrastive pairings ($[1,4], [2,5], [3,6]$), all-fragments ($[1,2,3,4,5,6]$),
and alternative pairing methods ($[1,2],[3,4],[5,6]$ and $[1,6],[2,5],[3,4]$) on IMDB-Clean-B with 40\% symmetric noise when $F=6$.
For feature extractor, all-fragments use a ResNet-34, while other pairing methods use ResNet-18 backbones.
}
\label{fig:contrasting_fragments}
\end{center}
\end{figure*}

\begin{figure*}[th]
\begin{center}
\centerline{\includegraphics[width=\textwidth]{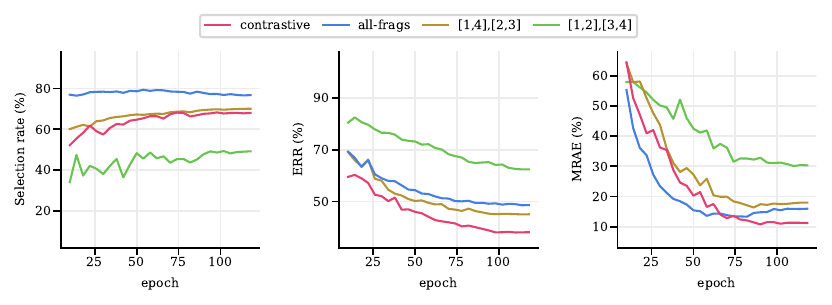}}
\vskip -0.2in
\caption{
\textbf{Fragment pairing analysis} compares contrastive pairings ($[1,3], [2,4]$), all-fragments ($[1,2,3,4]$),
and alternative pairing methods ($[1,2],[3,4]$ and $[1,4],[2,3]$) on IMDB-Clean-B with 40\% symmetric noise when $F=4$.
For feature extractor, all-fragments use a ResNet-34, while other pairing methods use ResNet-18 backbones.
}
\label{fig:contrasting_fragments_f4}
\end{center}
\end{figure*}

\begin{figure}[t]
    \begin{center}
    \centerline{\includegraphics[width=0.7\columnwidth]{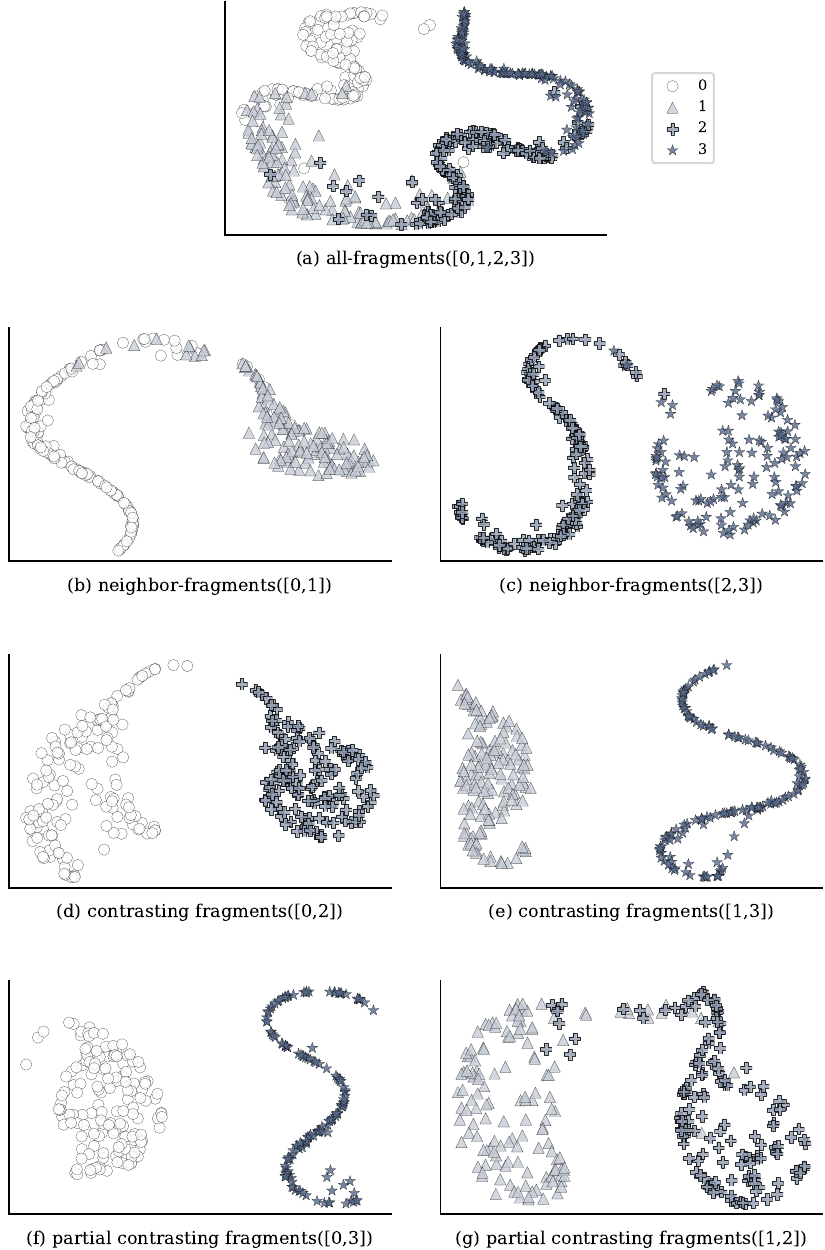}}
    \caption{\textbf{Detailed Representation Depiction}.
    A detailed comparison of the effect of fragment pairings via t-SNE visualization of the penultimate features from the feature extractors.
    The experiments are based on IMDB-Clean-B.
    }
    \label{fig:feature_space_depict}
    \end{center}
    \vskip -0.1in
\end{figure}

\subsection{Fragment Pairing Analysis}\label{subsec:contrast_combination}



In Fig.~\ref{fig:fragment_motivation}(c), we offer deeper insights into our approach by comparing contrastive fragment pairing ($[1,4], [2,5], [3,6]$) against all-fragments ($[1,2,3,4,5,6]$).
In Fig.~\ref{fig:discussion_main}(a), we show the importance of contrastive fragment pairing by comparing contrastive fragment pairing to alternative pairings.
In Fig.~\ref{fig:contrasting_fragments}--\ref{fig:contrasting_fragments_f4}, we present the extended results with Selection rate, ERR, and MRAE alongside other pairing methods. 

The experiments involve training the feature extractors using either contrastive fragment pairing, all-fragments, or alternative pairings.
Notably, a single feature extractor is employed for all fragments, whereas the fragment pairing (contrastive or alternative) uses a smaller feature extractor for each individual pair.
Subsequently, sample selection is executed in accordance with the Mixture of Neighboring Fragments approach (\S~\ref{subsec:mixture_of_contrasing_fragments}).

In an optimal selection algorithm, the Selection rate should approach $100 - \text{noise rate}(\%)$, with ERR and MRAE minimized.
Across all evaluation metrics, the contrastive fragment pairing demonstrates superior performance compared to other methods. 
It is important to highlight that performance is poorest when the pairing is least distinguishable ($[1,2],[3,4],[5,6]$ when $F=6$, $[1,2],[3,4]$ when $F=4$) and moderate when the pairing is partially distinguishable ($[1,6],[2,5],[3,4]$ when $F=6$, $[1,4], [2,3]$ when $F=4$).

Furthermore, in Fig.~\ref{fig:feature_space_depict}, we utilize t-SNE to compare the feature extractors trained using contrastive pairing, alternative pairings, and all-fragments. 
The visual comparison validates that representations trained with contrastive pairs exhibit significantly more distinguishable features.


\begin{figure*}[th]
\begin{center}
\centerline{\includegraphics[width=\textwidth]{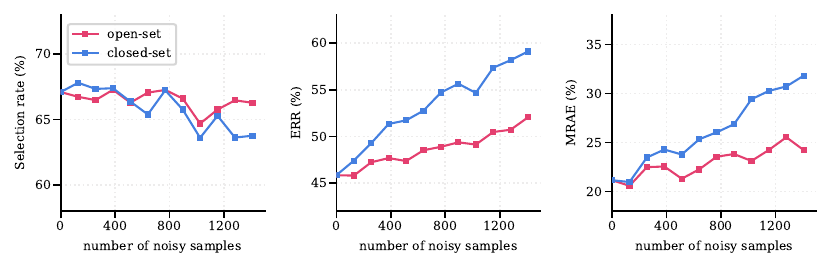}}
\caption{
\textbf{Closed-set/open-set noise analysis} displays the selection, ERR and MRAE
when closed-set or open-set noisy samples are injected into the clean dataset.
The experiments are based on IMDB-Clean-B.
}
\vskip -0.3in
\label{fig:disruptive_analysis}
\end{center}
\end{figure*}

\subsection{Closed-Set versus Open-Set Noise}\label{subsec:disruptive_anomaly_noise}
To explore the impact of closed-set and open-set noisy samples, as depicted in Fig.~\ref{fig:fragment_motivation}(b) in the main manuscript, we conducted an analysis of Selection rate, ERR, and MRAE performance while gradually introducing closed-set and open-set noisy samples into the IMDB-Clean-B dataset.
Our study employs the IMDB-Clean-B dataset, comprising a fixed set of clean samples that represent 40\% of the total dataset, alongside varying amounts of noisy samples. 
These noisy samples are classified into two distinct categories: closed-set and open-set noise \citep{wei2021open, wan2024unlocking}. 
For example, consider an example with 4 fragments, whose contrastive fragment pairs are $\{(1, 3), (2, 4)\}$.
When training a feature extractor on binary classification between fragment 1 and 3, noisy sample whose ground truth fragment id is either 2 or 4 but mislabeled as fragment id of either 1 or 3 is an open-set noisy sample.
On the other hand, a noisy sample whose ground truth fragment id is 1 but mislabeled as 3 (and vice versa) is a closed-set noisy sample.


Fig.~\ref{fig:disruptive_analysis} demonstrates that closed-set noisy samples have a considerably more adverse impact on ERR and MRAE compared to open-set noisy samples. 
Our contrastive fragment pair-based learning approach is advantageous in this regard, as it introduces open-set noisy samples in lieu of many closed-set noisy samples, thereby facilitating learning with reduced interference.


\subsection{Analysis of Samples on the Bounday versus Center of Fragments}\label{subsec:boundary_center}
Table~\ref{tab:anal_boundary_centre} presents a comparative analysis of the selection rate and error reduction rate (ERR) between samples located at the boundary and the center of fragments across eight experimental configurations. 
The results indicate an average difference of 2.29\% in selection rates and 2.43\% in ERR between the two groups. 
These findings substantiate the robustness of ConFrag's sample selection process, demonstrating consistent performance irrespective of the sample's positional location within the fragment.

\begin{table*}[t]
    \centering
    \caption{Difference of selection rate and ERR between the samples at the boundary and center of fragments}
    \begin{small}
    \begin{tabular}{lcccc}
    \toprule
    \multicolumn{5}{c}{\textbf{Selection rate}} \\
    \midrule
    \textbf{} & \textbf{20\%} & \textbf{40\%} & \textbf{60\%} & \textbf{80\%} \\
    \midrule
    \textbf{boundary} & 79.55\% & 65.31\% & 55.98\% & 65.96\% \\
    \textbf{center} & 80.18\% & 66.86\% & 54.64\% & 61.16\% \\
    \textbf{difference} & 0.63\% & 1.55\% & 1.34\% & 4.80\% \\
    \midrule
    \multicolumn{5}{c}{\textbf{ERR}} \\
    \midrule
    \textbf{boundary} & 31.90\% & 39.01\% & 55.44\% & 82.26\% \\
    \textbf{center} & 30.16\% & 36.63\% & 50.42\% & 82.91\% \\
    \textbf{difference} & 1.74\% & 2.38\% & 5.02\% & 0.65\% \\
    \bottomrule
    \end{tabular}
    \end{small}
    \label{tab:anal_boundary_centre}
\end{table*}

\begin{table}[t]
    \begin{center}
    \begin{small}
    \setlength{\tabcolsep}{4.2pt}
    \caption{\textbf{Ablation and Combination Analysis.} 
    The values are mean relative absolute error to the noise-free trained model on the IMDB-Clean-B~\citep{lin2021imdbclean} dataset, and lower values indicate better performances. 
    The results are the mean of three random seed experiments.
    }
    \label{tab:ablation}
    \begin{tabular}{lcccccc}
        \toprule
        \multicolumn{4}{c}{ablation and combinations}&\multicolumn{3}{c}{IMDB-Clean-B}
        \\\cmidrule(lr){1-4} \cmidrule(lr){5-7}
        & & & & \multicolumn{1}{c}{symmetric}    &\multicolumn{2}{c}{Gaussian} \\
        feat. ext. loss & backbone & jitter & Co-teaching & 40 & 30 & 50   \\
        \midrule
        CE  & ResNet-18 & &                        & 18.90 & 21.77 & 39.78  \\
        CE  & ResNet-18 & \checkmark &             & 12.64 & 15.70 & 33.36  \\
        CE  & ResNet-34 & \checkmark &             & 13.44 & 16.06 & 31.00  \\
        CE  & ResNet-18 & \checkmark &  \checkmark & 9.45 & 14.87 & 35.88  \\
        \midrule
        SCE & ResNet-18 &  &                        & 18.37 & 20.80 & 38.10  \\
        SCE & ResNet-18 & \checkmark &              & 16.84 & 20.07 & 38.18  \\
        SCE & ResNet-34 & \checkmark &              & 14.97 & 18.95 & 36.12   \\
        SCE & ResNet-18 & \checkmark &  \checkmark  & 13.19 & 18.32 & 41.02  \\
        \bottomrule
    \end{tabular}
    \end{small}
    \end{center}
\end{table}

\subsection{Ablation \& Combination Analysis}\label{subsec:ablation}
In Table~\ref{tab:ablation}, we present a comprehensive study comparing the performance of Cross-Entropy (CE) and Symmetric Cross Entropy (SCE)~\citep{wang19sce} losses in various ablation and combination experiments conducted on the IMDB-Clean-B dataset~\citep{lin2021imdbclean}, considering scenarios with 40\% symmetric noise and two variations of Gaussian random noise, each having a maximum standard deviation of 30 and 50.


Firstly, we illustrate the impact of jittering regularization through ablation on each of the losses.
Notably, jittering regularization emerges as a crucial component of ConFrag's performance, preventing the model from overfitting to the noisy labels.

The next ablation experiment entails replacing the ResNet-18 architecture of the feature extractors with ResNet-34.
The performance is enhanced when trained with SCE but decreases when trained with just CE.
This suggests that ConFrag could potentially benefit from a more powerful architecture, but it is not a necessity.

A significant advantage of ConFrag lies in its compatibility with other approaches. 
We showcase its performance when combined with an additional technique: Co-teaching~\citep{han18coteaching}, which is also employed by CNLCU and Co-Selfie in our baseline.
Co-teaching involves training the regression model while heuristically assuming that 25\% of the original noise still exists in the data (\eg 40\% original noise implies an assumption of 10\% noise during Co-teaching regression). 
Empirical observations reveal that Co-teaching consistently provides significant benefits. 

Upon comparing CE and SCE for feature extractor training loss, we observe that CE, when combined with jitter regularization, synergizes better to exhibit much stronger performance compared to SCE.

\begin{table}[t]
    \begin{center}
    \begin{small}
    \setlength{\tabcolsep}{4.2pt}
    \caption{\textbf{Discretized Baseline Analysis.} Mean Relative Absolute Error to the noise-free model of discretized versions of strongly performing models on the IMDB-Clean-B~\citep{lin2021imdbclean} dataset. 
    Lower is better. 
    }
    \label{tab:discrete}

    \begin{tabular}{lccc}
        \toprule
        &\multicolumn{3}{c}{IMDB-Clean-B}
        \\\cmidrule(lr){2-4}
        &\multicolumn{1}{c}{symmetric}    &\multicolumn{2}{c}{Gaussian} \\
        noise rate (\%)  & 40 & 30 & 50 \\
        \midrule
        CNLCU-S-D~\citep{xia22} & 55.71 & 64.71 & 79.59 \\
        CNLCU-S-D + mixup~\citep{xia22} & 55.14 & 67.17 & 81.32 \\
        CNLCU-H-D~\citep{xia22} & 37.76 & 51.36  & 76.40 \\
        CNLCU-H-D + mixup~\citep{xia22} & 65.32 & 67.31 & 84.22 \\
        Sigua-D~\citep{han20sigua} & 56.17 & 61.67 & 66.08 \\
        Sigua-D + mixup~\citep{han20sigua} & 33.55 & 29.33 & 49.44 \\
        BMM-D~\citep{arazo19} & 33.86 & 30.27 & 50.05 \\
        MD-DYR-SH-D~\citep{arazo19} & 33.89 & 31.18 & 51.23 \\
        CRUST-D~\citep{mirzasoleiman20crust} & 33.86 & 30.27 & 50.47\\
        CRUST-D + mixup~\citep{mirzasoleiman20crust} & 32.33 & 30.50 & 50.27 \\
        Selfie-D~\citep{song19b} & 31.50 & 24.86 & 47.46 \\
        Selfie-D + mixup~\citep{song19b} & 35.33 & 28.02  &  46.42 \\
        Co-Selfie-D~\citep{song19b} & 30.20 & 26.36 & 49.61 \\
        Co-Selfie-D + mixup~\citep{song19b} & 33.18 & 28.28 & 52.20 \\
        \specialrule{0.7pt}{1pt}{1pt}
        ConFrag  (Ours) & 12.64 & 15.70 & 33.36 \\
        Co-ConFrag (Ours) & 9.45 & 14.87  & 35.88 \\
        \bottomrule
    \end{tabular}
    \end{small}
    \end{center}
\end{table}

\subsection{Discretized Baselines}\label{subsec:discrete_baselines}
In Table~\ref{tab:discrete}, we present a discretized version of several strong baselines, including Sigua~\citep{han20sigua}, CNLCU~\citep{xia22}, BMM~\citep{arazo19}, Selfie/Co-Selfie~\citep{song19b}, MD-DYR-SH~\citep{arazo19}, and CRUST~\citep{mirzasoleiman20crust}.

The discretization process aligns with our fragmentation approach used for ConFrag. We obtain selected samples at the end of every epoch to independently train the regression model. 
Additionally, we report performance with mixup~\citep{zhang18mixup}, a technique that proves beneficial for some baselines like Sigua~\citep{han20sigua}.

Notably, most baselines exhibit a deterioration in performance following discretization. 
However, Selfie/Co-Selfie~\citep{song19b} stands out as the exception, showing an improvement in performance after discretization. 
Interestingly, Sigua is the sole method that benefits from mixup~\citep{zhang18mixup} training.

\begin{figure}[t]
    \begin{center}
    \centerline{\includegraphics[width=\columnwidth]{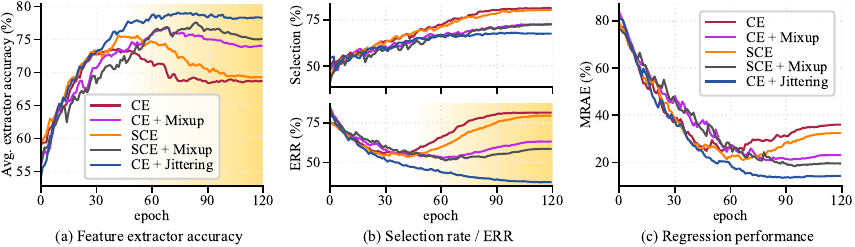}}
    \caption{\textbf{Comparison of regularization methods}.
    Compared to other regularization methods, neighborhood jittering demonstrates superior performance in 
    (a) feature extractor test accuracy, (b) ERR, and (c) performance in regression.
    The analysis is conducted on IMDB-Clean-B with symmetric 40\% noise.
    }
    \label{fig:jitter_comparison}
    \end{center}
\end{figure}

\subsection{Comparison of Neighborhood Jittering and Other Regularization Methods}\label{subsec:jitter_comparison}
In Fig.~\ref{fig:jitter_comparison}, we compare neighborhood jittering with other regularization methods 
that can be applied to classification-based feature extractors 
(SCE with weight decay~\citep{wang19sce}, mixup~\citep{zhang18mixup}, and their combinations).
In conclusion, neighborhood jittering exhibits the strongest performance in 
feature extractor test accuracy, ERR, and MRAE, among other regularization methods. 
It is observed that ERR and MRAE improve in line with the performance of the feature extractor.

\subsection{Extended Selection Rate/ERR/MRAE Comparison and Analysis}\label{subsec:ERR}
\begin{figure*}[th]
\begin{center}
\centerline{\includegraphics[width=0.75\textwidth]{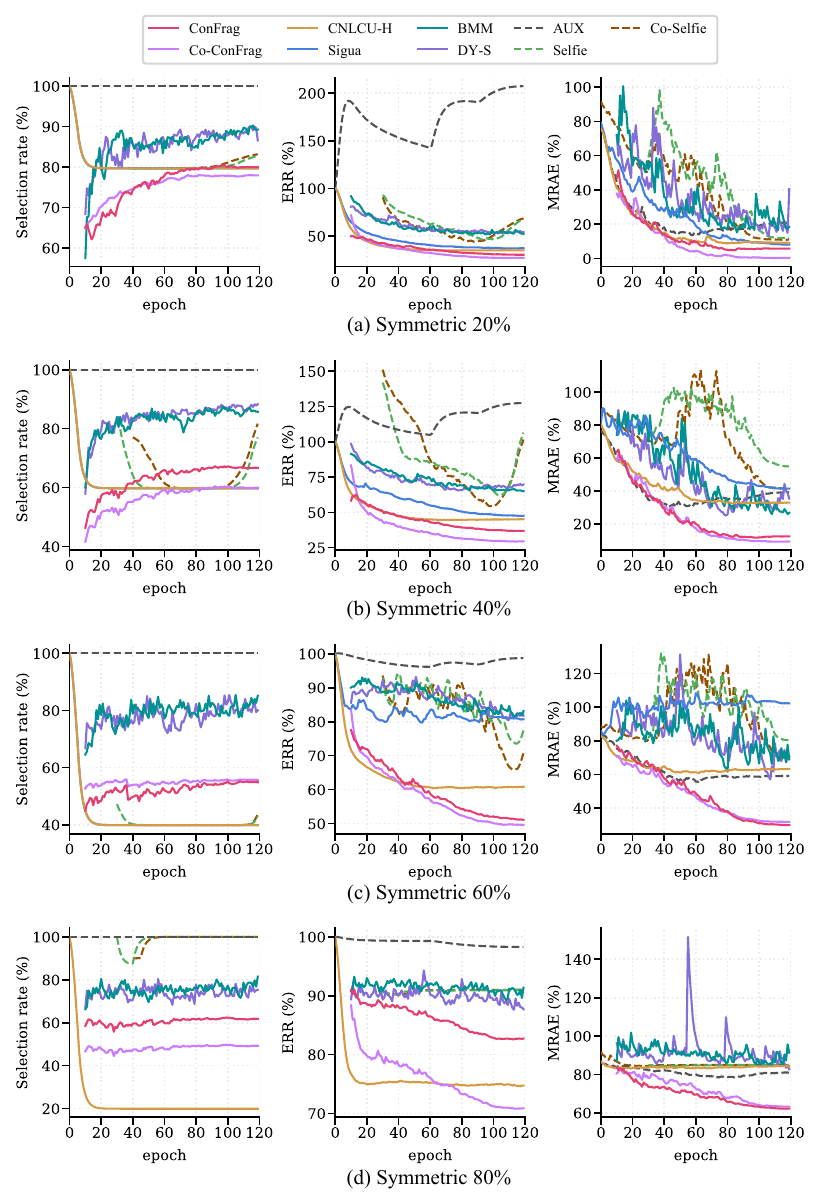}}
\caption{\textbf{Selection, ERR and MRAE comparison} of ConFrag, Co-ConFrag and filtering/refurbishment baselines on IMDB-Clean-B
with symmetric 20\%(a), 40\%(b), 60\%(c) and 80\%(d) noise, repectively.
}
\label{fig:selerr_comparison_supp1}
\end{center}
\end{figure*}

\begin{figure*}[th]
\begin{center}
\centerline{\includegraphics[width=0.75\textwidth]{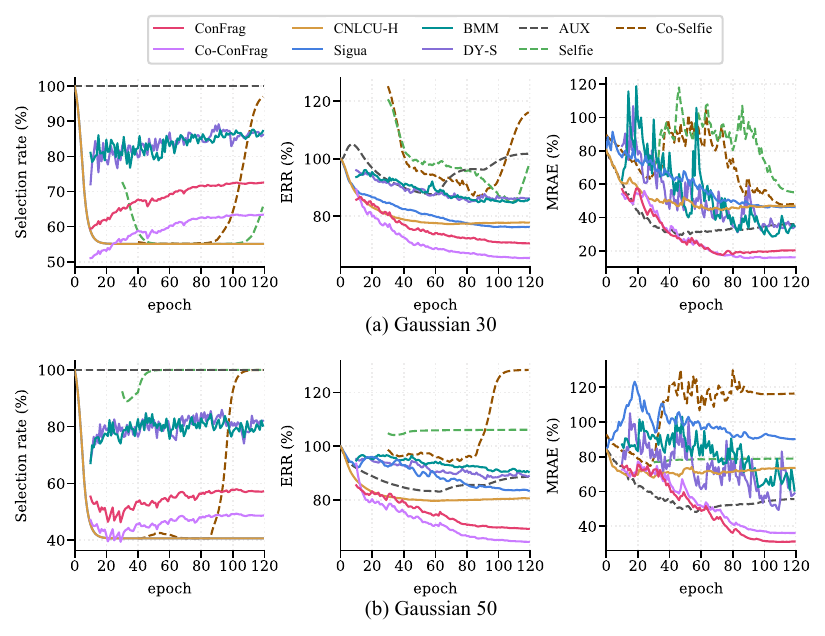}}
\caption{\textbf{Selection, ERR and MRAE comparison} of ConFrag, Co-ConFrag and filtering/refurbishment baselines on IMDB-Clean-B
with Gaussian 30(a) and Gaussian 50(b) noise, repectively.
}
\label{fig:selerr_comparison_supp2}
\end{center}
\end{figure*}

In addition to presenting the Selection rate, ERR and MRAE
for symmetric 40\%, Gaussian 30, and Gaussian 50 noise experiments on the IMDB-Clean-B dataset in the main manuscript,
we have included results for all noise types, along with additional baselines (CNLCU-H, Sigua, BMM, DY-S, AUX, Selfie, Coselfie),
in both Fig.~\ref{fig:selerr_comparison_supp1} and Fig.~\ref{fig:selerr_comparison_supp2}.

As mentioned in \S~\ref{subsec:evaluation_metrics}, the ideal scenario for selection and refurbishment methods involves achieving a high selection rate while maintaining a low ERR, 
resulting in a reduced MRAE. We examine the relationship between the selection rate, ERR, and MRAE based on Fig.~\ref{fig:selerr_comparison_supp1}(b). 
As training progresses, ConFrag and other selection methods (CNLCU-H, Sigua, BMM, DY-S) approach the ideal condition, resulting in an improving trend in MRAE. 
ConFrag, in particular, comes closest to the ideal scenario, resulting in superior MRAE performance.

The most unfavorable scenario arises when there is a low selection rate coupled with a high ERR. 
Selfie exemplifies the scenario in Fig.~\ref{fig:selerr_comparison_supp1}(b), which is connected to a relatively worse MRAE.

The scenarios of the low selection rates with low ERR and the high selection rates with high ERR can be further examined using CNLCU-H and BMM. 
CNLCU-H demonstrates superior selection quality in terms of ERR, while BMM exhibits a higher quantity in the selection rate. 
This quality/quantity trade-off is linked to the observation that CNLCU-H and BMM show similar MRAE performance in Fig.~\ref{fig:selerr_comparison_supp1}(b). 
Additionally, Fig.~\ref{fig:selerr_comparison_supp2}(a) reveals that the selection rate gap widens, while the ERR gap narrows when compared to Fig.~\ref{fig:selerr_comparison_supp1}(b). 
This is associated with BMM outperforming CNLCU-H in terms of the MRAE.

It's important to note that, rather than employing the selection rate and ERR as indicators for MRAE, 
these metrics offer valuable insights when assessing selected or refurbished samples 
directly independent of any potential regularizing effects introduced by the underlying regression model.

In addition, upon a detailed analysis of the figures, it becomes evident that Co-ConFrag consistently achieves the lowest ERR across a wide range of noise types. 
Notably, it maintains a Selection rate of above 40\% while maintaining low ERR even in the presence of severe noise conditions, which leads to outstanding MRAE performance.



\subsection{Variance Across Random Seeds}\label{subsec:variance}
\begin{figure*}[th]
\begin{center}
\centerline{\includegraphics[width=0.75\textwidth]{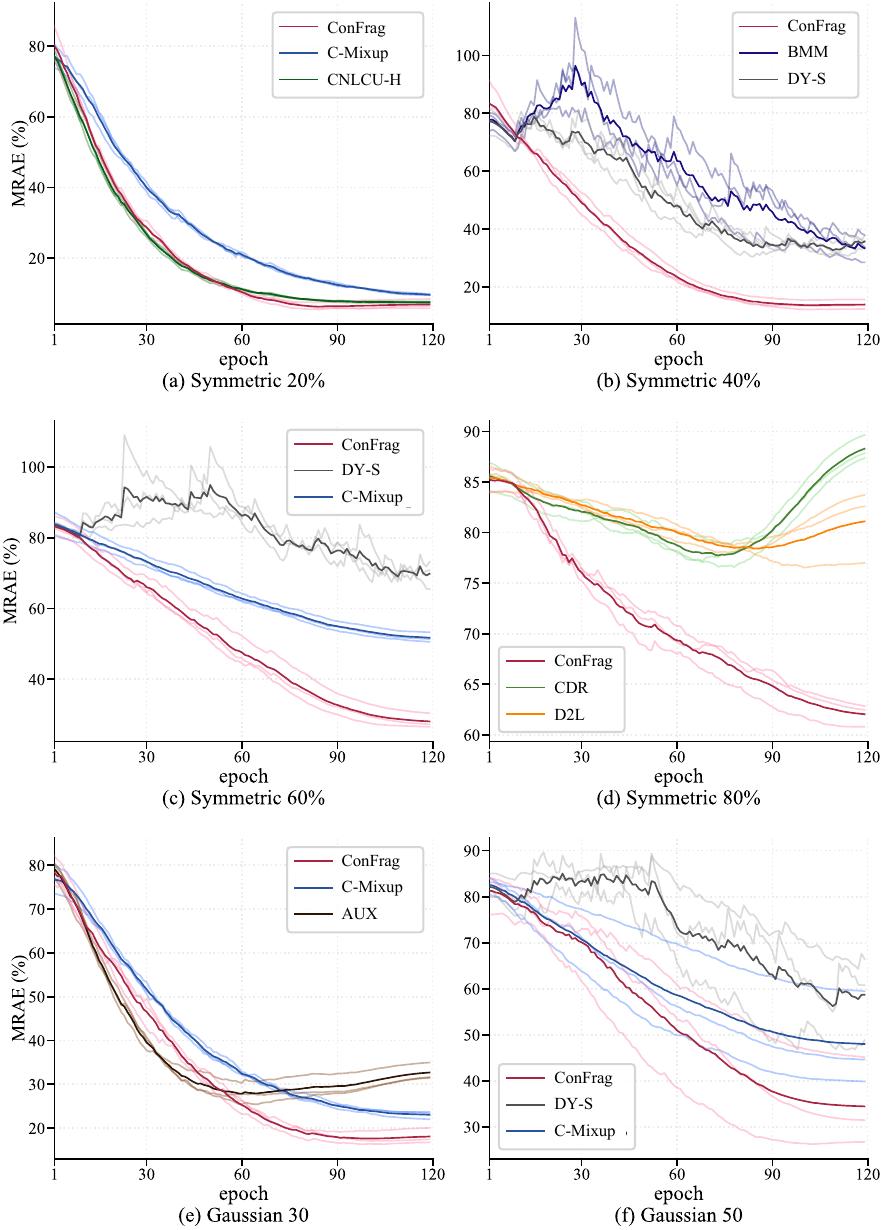}}
\caption{\textbf{Variance Analysis} of three unique random seed experiments on IMDB-Clean-B.  
The top two best-performing baselines under each noise type are reported.
}
\label{fig:variance_analysis}
\end{center}
\vskip -0.2in
\end{figure*}

In Fig.~\ref{fig:variance_analysis}, we plot the variance of three unique random seed experiments on all six noise types (symmetric 20\%/40\%/60\%/80\%, Gaussian 30/50) 
on the IMDB-Clean-B dataset. To declutter the graph, we compare it against the top two best-performing baselines under each noise type.

Tab.~\ref{tab:mrae_with_std}--\ref{tab:mrae_with_std_2} show the main experimental results of Tab.~\ref{tab:main_mrae} with standard deviation.

\subsection{Standard Mean Absolute Error}\label{subsec:mae}
In Tables~\ref{tab:main_mae}--\ref{tab:main_mae_2}, we report the standard mean absolute error (along with standard deviation) within the respective label ranges for each dataset.

\clearpage

\begin{table*}[t]
    \caption{\textbf{Mean Relative Absolute Error (\%)} and its standard deviation to the noise-free trained model on the AFAD-B, IMDB-Clean-B and IMDB-WIKI-B datasets.
    Lower is better. 
    A negative value indicates it performs even better than the noise-free model.
    The results are the mean of three random seed experiments.
    Number in parenthesis indicates standard deviation.
    The best and the second best methods are respectively marked in \textcolor{red}{red} and \textcolor{blue}{blue}.
    CNLCU-S/H, Co-Selfie, and Co-ConFrag use dual networks to teach each other as done in \citet{han18coteaching}.
    }
    \begin{center}
    \begin{small}
    \setlength{\tabcolsep}{4.2pt}
    \resizebox{\columnwidth}{!}{%
    \begin{tabular}{lccccccccccccc}
        \toprule
        &\multicolumn{6}{c}{AFAD-B}         &\multicolumn{6}{c}{IMDB-Clean-B} & IMDB-WIKI-B
        \\\cmidrule(lr){2-7}\cmidrule(lr){8-13}\cmidrule(lr){14-14}
        &\multicolumn{4}{c}{symmetric}    &\multicolumn{2}{c}{Gaussian} &\multicolumn{4}{c}{symmetric} &\multicolumn{2}{c}{Gaussian} & real noise
        \\\cmidrule(lr){2-5}\cmidrule(lr){6-7}\cmidrule(lr){8-11}\cmidrule(lr){12-13}\cmidrule(lr){14-14}
        noise rate  & 20 & 40 & 60 & 80 & 30 & 50 & 20 & 40 & 60 & 80 & 30 & 50 & - \\
        \midrule
        \multirow{2}{*}{Vanilla} & 9.37  & 20.27 & 30.65 & 43.09 & 28.77 & 39.03 & 16.18 & 32.05 & 53.13 & 76.35 & 26.89 & 50.28 & 00.00 \\
            & (0.72) & (0.93) & (1.15) & (45.96) & (1.57) & (4.32) & (1.60) & (0.20) & (0.93) & (1.29) & (2.45) & (9.07) & (00.00)\\
        \specialrule{0.1pt}{1pt}{1pt}
        \multirow{2}{*}{CNLCU-S} & 10.98 & 20.44 & 32.44 & 41.99 & 30.60 & 40.66 & 51.40 & 66.62 & 82.83 & 85.65 & 83.39 & 82.10 & 21.54 \\
            & (0.42) & (3.60) & (0.18) & (46.23) & (1.11) & (5.33) & (2.03) & (2.74) & (2.82) & (0.86) & (10.54) & (4.38) & (1.28)\\
        \multirow{2}{*}{CNLCU-H} & 4.63 & 16.32 & 36.01 & 44.71 & 35.68 & 43.64 & 6.84 & 31.16 & 63.08 & 82.65 & 46.53 & 65.24 & -2.93 \\
            & (0.77) & (1.51) & (3.39) & (28.95) & (3.08) & (2.79) & (0.64) & (1.29) & (2.01) & (1.65) & (5.60) & (7.16) & (0.82)\\
        \multirow{2}{*}{Sigua} & 5.96 & 21.09 & 43.33 & 49.71 & 42.52 & 46.19 & 9.82 & 46.17 & 77.59 & 85.62 & 60.97 & 77.42 & 1.96 \\
            & (1.43) & (2.15) & (2.12) & (53.69) & (3.47) & (3.85) & (0.54) & (9.52) & (2.01) & (0.94) & (19.19) & (2.07) & (1.65)\\
        \multirow{2}{*}{SPR} & 9.74 & 18.85 & 30.43 & 43.25 & 28.50 & 39.69 & 14.47 & 32.44 & 54.88 & 79.37 & 25.67 & 51.05 & -0.93 \\
            & (0.53) & (1.27) & (0.47) & (43.74) & (1.24) & (6.72) & (1.06) & (0.49) & (1.76) & (1.30) & (1.12) & (10.31) & (1.61)\\
        \multirow{2}{*}{BMM} & 5.60 & 15.00 & 39.15 & 46.41 & 30.96 & 44.00 & 8.85 & 21.54 & 55.57 & 80.40 & 24.33 & 57.21 & 17.88 \\
            & (0.68) & (1.91) & (3.16) & (44.77) & (6.89) & (2.11) & (1.53) & (2.29) & (8.88) & (3.18) & (1.49) & (16.74) & (2.05)\\
        \multirow{2}{*}{DY-S} & 6.87 & 15.56 & 32.24 & 45.72 & 24.40 & 43.41 & 10.42 & 21.90 & 49.94 & 78.16 & 24.70 & 44.56 & -3.41 \\
            & (2.22) & (3.26) & (5.14) & (33.46) & (0.68) & (4.87) & (0.96) & (1.58) & (0.26) & (0.77) & (2.23) & (10.04) & (0.86)\\
        \multirow{2}{*}{C-Mixup} & \textcolor{blue}{2.74} & 14.80 & 27.17 & 41.95 & 24.28 & 36.91 & 8.82 & 27.74 & 50.87 & 76.79 & 21.92 & 47.04 & \textcolor{blue}{-5.26} \\
            & (0.74) & (0.16) & (0.77) & (38.53) & (2.29) & (7.88) & (0.25) & (0.46) & (1.28) & (0.86) & (0.96) & (10.33) & (0.52)\\
        \multirow{2}{*}{RDI} & 10.64 & 21.80 & 39.32 & 47.07 & 37.33 & 44.41 & 16.35 & 29.33 & 55.91 & 79.92 & 25.69 & 51.35 & 1.06 \\
            & (0.41) & (0.66) & (0.76) & (49.28) & (1.51) & (4.41) & (1.09) & (1.98) & (0.60) & (0.69) & (1.75) & (12.37) & (0.67)\\
        \multirow{2}{*}{CDR} & 10.26 & 18.71 & 32.27 & 43.38 & 29.74 & 39.21 & 17.47 & 32.19 & 54.75 & 75.45 & 28.46 & 51.73 & -0.39 \\
            & (1.20) & (1.03) & (0.66) & (45.06) & (0.41) & (6.15) & (1.11) & (1.54) & (1.72) & (0.89) & (2.89) & (7.38) & (1.28)\\
        \multirow{2}{*}{D2L} & 9.43 & 20.75 & 31.25 & 44.50 & 28.86 & 40.10 & 16.94 & 33.85 & 55.54 & 76.28 & 29.30 & 52.44 & -0.66 \\
            & (0.41) & (2.51) & (0.29) & (45.37) & (1.01) & (5.73) & (1.34) & (2.21) & (0.96) & (1.00) & (3.73) & (9.20) & (0.82)\\
        \multirow{2}{*}{AUX} & 6.15 & 19.01 & 31.16 & 42.83 & 28.28 & 39.05 & 12.58 & 28.82 & 52.33 & 76.75 & 23.27 & 49.42 & -3.67 \\
            & (0.17) & (0.71) & (0.50) & (44.84) & (1.70) & (4.81) & (0.66) & (1.35) & (0.82) & (1.08) & (1.78) & (9.86) & (0.72)\\
        \multirow{2}{*}{Selfie} & 16.91 & 25.02 & 44.18 & 47.78 & 46.02 & 50.73 & 27.43 & 53.74 & 79.38 & 84.00 & 60.68 & 78.03 & 14.00 \\
            & (4.09) & (3.42) & (0.48) & (42.55) & (5.90) & (4.93) & (15.12) & (2.07) & (0.08) & (0.28) & (5.42) & (3.46) & (11.45)\\
        \multirow{2}{*}{Co-Selfie} & 14.61 & 22.95 & 39.79 & 47.72 & 41.05 & 53.00 & 23.52 & 50.07 & 67.42 & 84.25 & 52.44 & 74.73 & -0.44 \\
            & (2.66) & (1.03) & (1.33) & (35.29) & (6.05) & (15.38) & (12.18) & (11.39) & (3.08) & (0.59) & (8.15) & (6.99) & (5.19))\\
        \multirow{2}{*}{Superloss} & 7.36 & 18.24 & 29.78 & 44.26 & 27.59 & 42.96 & 23.38 & 45.41 & 67.11 & 80.85 & 53.88 & 63.33 & -3.58 \\
            & (2.02) & (1.38) & (1.64) & (40.38) & (1.09) & (5.80) & (4.49) & (3.14) & (1.88) & (1.66) & (16.22) & (8.88) & (1.46)\\
        \specialrule{0.7pt}{1pt}{1pt}
        \multirow{2}{*}{\textbf{ConFrag}}  & \textcolor{blue}{2.74} & \textcolor{blue}{8.16} & \textcolor{red}{15.91} & \textcolor{blue}{34.42} & \textcolor{blue}{17.49} & \textcolor{red}{27.31} & \textcolor{blue}{5.08} & \textcolor{blue}{12.64} & \textcolor{red}{27.26} & \textcolor{red}{61.24} & \textcolor{blue}{15.70} & \textcolor{red}{33.36} & -3.06 \\
            & (0.95) & (0.43) & (0.39) & (22.14) & (1.12) & (5.31) & (0.36) & (1.94) & (2.25) & (1.42) & (1.43) & (10.14) & (1.25)\\
        \multirow{2}{*}{\textbf{Co-ConFrag}} & \textcolor{red}{0.54} & \textcolor{red}{7.25} & \textcolor{blue}{16.65} & \textcolor{red}{33.93} & \textcolor{red}{17.43} & \textcolor{blue}{28.26} & \textcolor{red}{1.50} & \textcolor{red}{9.45} & \textcolor{blue}{28.44} & \textcolor{blue}{61.36} & \textcolor{red}{14.87} & \textcolor{blue}{35.88} & \textcolor{red}{-8.86} \\
            & (0.77) & (0.59) & (0.14) & (18.48) & (0.99) & (2.45) & (0.71) & (0.62) & (3.09) & (3.14) & (0.23) & (11.44) & (0.83)\\
        \bottomrule
    \end{tabular}
    }
    \end{small}
    \end{center}
    \label{tab:mrae_with_std}
    \vskip -0.2in
\end{table*}

\begin{table*}[t]
    \caption{\textbf{Mean Relative Absolute Error (\%)} and its standard deviation to the noise-free trained model on the SHIFT15M-B and MSD-B datasets.
    Lower is better. A negative value indicates it performs even better than the noise-free model.
    The results are the mean of three random seed experiments.
    Number in parenthesis indicates standard deviation.
    The best and the second best methods are respectively marked in \textcolor{red}{red} and \textcolor{blue}{blue}.
    CNLCU-S/H, Co-Selfie, and Co-ConFrag use dual networks to teach each other as done in \citet{han18coteaching}.
    SPR~\citep{wang22spr} fails to run for SHIFT15M-B due to excessive memory consumption.}
    \begin{center}
    \begin{small}
    \setlength{\tabcolsep}{4.2pt}
    \resizebox{\columnwidth}{!}{%
    \begin{tabular}{lcccccccccccc}
        \toprule
        &\multicolumn{6}{c}{SHIFT15M-B}         &\multicolumn{6}{c}{MSD-B}
        \\\cmidrule(lr){2-7}\cmidrule(lr){8-13}
        &\multicolumn{4}{c}{symmetric}    &\multicolumn{2}{c}{Gaussian} &\multicolumn{4}{c}{symmetric} &\multicolumn{2}{c}{Gaussian}
        \\\cmidrule(lr){2-5}\cmidrule(lr){6-7}\cmidrule(lr){8-11}\cmidrule(lr){12-13} 
        noise rate & 20 & 40 & 60 & 80 & 30 & 50 & 20 & 40 & 60 & 80 & 30 & 50 \\
        \midrule
        \multirow{2}{*}{Vanilla}            & 9.11 & 17.96 & 27.02 & 36.34 & 6.54 & 15.16 & 8.23 & 18.43 & 31.67 & 45.85 & 6.96 & 15.74 \\
            & (0.56) & (1.50) & (0.96) & (0.08) & (1.11) & (0.90) &	(0.18) & (2.47) & (3.51) & (0.36) & (0.65) & (3.03) \\
        \specialrule{0.1pt}{1pt}{1pt}
        \multirow{2}{*}{CNLCU-S} & 12.98 & 19.42 & 24.31 & 34.47 & 15.33 & 20.90 & 0.13 & 6.04 & 21.52 & 46.01 & 4.75 & 12.51 \\
            & (0.15) & (0.39) & (0.70) & (0.17) & (1.09) & (0.34) &	(1.18) & (0.31) & (3.61) & (1.51) & (0.92) & (1.08) \\
        \multirow{2}{*}{CNLCU-H} & 6.26 & 12.84 & 20.04 & 36.03 & 8.88 & 15.65 & 0.27 & 4.98 & 10.32 & 29.83 & 5.11 & 9.22 \\
            & (0.43) & (0.55) & (0.42) & (0.80) & (0.36) & (0.19) &	(0.19) & (0.94) & (1.46) & (1.33) & (0.31) & (0.39) \\
        \multirow{2}{*}{Sigua} & 6.94 & 14.09 & 26.08 & 37.03 & 10.32 & 17.44 & 1.29 & 7.19 & 17.35 & 50.87 & 6.80 & 12.38 \\
            & (0.13) & (0.17) & (0.29) & (2.84) & (0.77) & (0.70) &	(0.21) & (0.64) & (3.30) & (3.33) & (1.44) & (0.10) \\
        \multirow{2}{*}{SPR} &-&-&-&-&-&-& 7.07 & 18.19 & 33.39 & 45.61 & 5.01 & 15.36 \\
           & (-) & (-) & (-) & (-) & (-) & (-) & (1.30) & (1.32) & (3.17) & (2.11) & (0.31) & (2.15) \\
        \multirow{2}{*}{BMM} & 6.96 & 12.42 & 18.64 & 26.79 & 7.58 & 13.13 & 3.32 & 10.30 & 23.40 & 43.56 & 5.29 & 11.85 \\
            & (0.52) & (1.16) & (0.78) & (1.04) & (1.24) & (0.55) &	(0.64) & (1.85) & (2.31) & (1.95) & (0.63) & (0.77) \\
        \multirow{2}{*}{DY-S} & 7.11 & 11.94 & 18.85 & 29.04 & 6.90 & 13.50 & 3.39 & 8.06 & 18.65 & 35.24 & 4.77 & 9.83 \\
            & (0.25) & (0.74) & (0.10) & (1.48) & (0.97) & (0.95) &	(1.07) & (1.19) & (3.50) & (1.83) & (1.28) & (1.33) \\
        \multirow{2}{*}{C-Mixup} & 9.47 & 16.15 & 24.08 & 34.17 & 5.88 & 14.51 & 3.75 & 13.13 & 26.73 & 40.90 & 2.96 & 10.97 \\
            & (0.23) & (0.83) & (0.32) & (0.40) & (1.03) & (0.98) &	(0.83) & (2.34) & (2.16) & (2.07) & (0.17) & (0.38) \\
        \multirow{2}{*}{RDI} & 9.91 & 17.92 & 26.63 & 36.29 & 7.08 & 15.18 & 21.04 & 30.09 & 38.78 & 49.49 & 19.19 & 27.88 \\
            & (0.45) & (0.27) & (0.36) & (0.67) & (0.85) & (0.84) &	(0.09) & (0.22) & (0.98) & (1.12) & (0.69) & (1.60) \\
        \multirow{2}{*}{CDR} & 9.52 & 17.78 & 26.97 & 35.97 & 7.14 & 15.17 & 7.83 & 17.86 & 32.83 & 45.91 & 6.73 & 16.92 \\
            & (1.17) & (0.83) & (0.36) & (0.73) & (1.01) & (0.72) &	(0.91) & (3.23) & (2.36) & (0.74) & (0.49) & (1.55) \\
        \multirow{2}{*}{D2L} & 9.25 & 18.03 & 26.55 & 36.23 & 6.34 & 15.60 & 7.13 & 19.96 & 32.47 & 46.64 & 5.51 & 15.54 \\
            & (0.26) & (1.47) & (1.13) & (0.66) & (0.69) & (1.58) &	(0.37) & (1.08) & (2.21) & (2.56) & (0.76) & (2.05) \\
        \multirow{2}{*}{AUX} & 7.74 & 16.95 & 26.61 & 36.47 & 4.92 & 14.40 & 6.12 & 18.18 & 31.09 & 45.70 & 5.21 & 15.45 \\
            & (0.33) & (1.03) & (0.30) & (0.50) & (1.11) & (0.94) &	(0.88) & (1.55) & (3.07) & (1.43) & (0.28) & (1.78) \\
        \multirow{2}{*}{Selfie} & 4.84 & 10.22 & 22.28 & 38.15 & 5.51 & 11.58 & 1.43 & 8.40 & 20.24 & 45.87 & 14.37 & 24.13 \\
            & (0.77) & (0.71) & (2.82) & (0.42) & (0.97) & (0.45) &	(0.24) & (1.30) & (4.61) & (2.88) & (3.28) & (3.41) \\
        \multirow{2}{*}{Co-Selfie} & 11.53 & 16.43 & 32.08 & 39.32 & 13.45 & 22.33 & \textcolor{blue}{-0.38} & \textcolor{blue}{4.41} & \textcolor{red}{8.32} & 35.47 & 6.78 & 13.15 \\
            & (0.84) & (0.62) & (0.64) & (0.54) & (0.74) & (0.85) &	(0.12) & (0.68) & (1.40) & (0.57) & (1.70) & (1.60) \\
        \multirow{2}{*}{Superloss} & 5.44 & 12.26 & 23.23 & 35.24 & 5.60 & 13.28 & -0.15 & 10.68 & 23.15 & 45.55 & 4.35 & 16.36 \\
            & (1.03) & (1.48) & (1.89) & (0.28) & (1.28) & (0.67) &	(0.29) & (2.10) & (3.15) & (6.77) & (0.74) & (2.99) \\
        \specialrule{0.7pt}{1pt}{1pt}
        \multirow{2}{*}{\textbf{ConFrag}} & \textcolor{blue}{2.46} & \textcolor{blue}{6.18} & \textcolor{red}{10.68} & \textcolor{blue}{19.04} & \textcolor{blue}{3.66} & \textcolor{red}{8.09} & 0.57 & 4.94 & 11.22 & \textcolor{blue}{23.41} & 2.39 & \textcolor{blue}{6.49} \\
            & (0.42) & (0.45) & (0.65) & (0.63) & (0.37) & (0.05) &	(0.43) & (0.34) & (1.38) & (2.00) & (0.84) & (1.90) \\
        \multirow{2}{*}{\textbf{Co-ConFrag}} & \textcolor{red}{0.85} & \textcolor{red}{5.52} & \textcolor{blue}{10.80} & \textcolor{red}{18.83} & \textcolor{red}{3.03} & \textcolor{blue}{8.70} & \textcolor{red}{-0.65} & \textcolor{red}{2.98} & \textcolor{blue}{8.66} & \textcolor{red}{20.53} & \textcolor{red}{1.73} & \textcolor{red}{6.00} \\
           & (0.31) & (0.66) & (0.43) & (0.41) & (0.94) & (0.46) &	(0.72) & (0.66) & (0.36) & (2.46) & (1.02) & (1.07) \\
        \bottomrule
    \end{tabular}
    }
    \end{small}
    \end{center}
    \label{tab:mrae_with_std_2}
    \vskip -0.2in
\end{table*}

\begin{table*}[t]
    \caption{\textbf{Standard Mean Absolute Error} and its standard deviation to the noise-free trained model on the AFAD-B, IMDB-Clean-B and IMDB-WIKI-B datasets.
    Lower is better. 
    The results are the mean of three random seed experiments.
    Number in parenthesis indicates standard deviation.
    The best and the second best methods are respectively marked in \textcolor{red}{red} and \textcolor{blue}{blue}.
    CNLCU-S/H, Co-Selfie, and Co-ConFrag use dual networks to teach each other as done in \citet{han18coteaching}.
    }
    \begin{center}
    \begin{small}
    \setlength{\tabcolsep}{4.2pt}
    \resizebox{\columnwidth}{!}{%
    \begin{tabular}{lccccccccccccc}
        \toprule
        &\multicolumn{6}{c}{AFAD-B}         &\multicolumn{6}{c}{IMDB-Clean-B} & IMDB-WIKI-B
        \\\cmidrule(lr){2-7}\cmidrule(lr){8-13}\cmidrule(lr){14-14}
        &\multicolumn{4}{c}{symmetric}    &\multicolumn{2}{c}{Gaussian} &\multicolumn{4}{c}{symmetric} &\multicolumn{2}{c}{Gaussian} & real noise
        \\\cmidrule(lr){2-5}\cmidrule(lr){6-7}\cmidrule(lr){8-11}\cmidrule(lr){12-13}\cmidrule(lr){14-14}
        noise rate  & 20 & 40 & 60 & 80 & 30 & 50 & 20 & 40 & 60 & 80 & 30 & 50 & - \\
        \midrule
        \multirow{2}{*}{Vanilla}            & 4.75 & 5.22 & 5.68 & 6.22 & 5.59 & 6.04 & 8.11 & 9.22 & 10.70 & 12.32 & 8.86 & 10.50 & 7.23 \\
            & (0.02) & (0.03) & (0.05) & (0.06) & (0.08) & (0.18) &	(0.09) & (0.05) & (0.12) & (0.06) & (0.13) & (0.60) & (0.09)\\
        \specialrule{0.1pt}{1pt}{1pt}
        \multirow{2}{*}{CNLCU-S} & 4.82 & 5.23 & 5.75 & 6.17 & 5.67 & 6.11 & 10.57 & 11.64 & 12.77 & 12.97 & 12.81 & 12.72 & 8.78\\
            & (0.03) & (0.14) & (0.01) & (0.09) & (0.07) & (0.22) &	(0.15) & (0.20) & (0.15) & (0.04) & (0.78) & (0.25) & (0.16)\\
        \multirow{2}{*}{CNLCU-H} & 4.55 & 5.05 & 5.91 & 6.29 & 5.89 & 6.24 & 7.46 & 9.16 & 11.39 & 12.76 & 10.24 & 11.54 & 7.01\\
            & (0.05) & (0.05) & (0.16) & (0.08) & (0.15) & (0.11) &	(0.08) & (0.12) & (0.14) & (0.18) & (0.43) & (0.47) & (0.02)\\
        \multirow{2}{*}{Sigua} & 4.60 & 5.26 & 6.23 & 6.50 & 6.19 & 6.35 & 7.67 & 10.21 & 12.40 & 12.96 & 11.25 & 12.39 & 7.37\\
            & (0.07) & (0.10) & (0.07) & (0.04) & (0.17) & (0.17) &	(0.04) & (0.67) & (0.08) & (0.04) & (1.37) & (0.19) & (0.12)\\
        \multirow{2}{*}{SPR} & 4.77 & 5.16 & 5.67 & 6.22 & 5.58 & 6.07 & 8.00 & 9.25 & 10.82 & 12.53 & 8.78 & 10.55 & 7.16\\
            & (0.04) & (0.04) & (0.03) & (0.06) & (0.07) & (0.28) &	(0.09) & (0.02) & (0.08) & (0.12) & (0.05) & (0.70) & (0.03)\\
        \multirow{2}{*}{BMM} & 4.59 & 5.00 & 6.04 & 6.36 & 5.69 & 6.26 & 7.60 & 8.49 & 10.87 & 12.60 & 8.68 & 10.98 & 8.52\\
            & (0.03) & (0.09) & (0.12) & (0.06) & (0.31) & (0.09) &	(0.11) & (0.20) & (0.68) & (0.21) & (0.12) & (1.16) & (0.13)\\
        \multirow{2}{*}{DY-S} & 4.64 & 5.02 & 5.74 & 6.33 & 5.40 & 6.23 & 7.71 & 8.51 & 10.47 & 12.44 & 8.71 & 10.10 & 6.98\\
            & (0.09) & (0.12) & (0.24) & (0.16) & (0.05) & (0.20) &	(0.11) & (0.13) & (0.04) & (0.05) & (0.14) & (0.68) & (0.07)\\
        \multirow{2}{*}{C-Mixup} & \textcolor{blue}{4.46} & 4.99 & 5.52 & 6.17 & 5.40 & 5.95 & 7.60 & 8.92 & 10.54 & 12.35 & 8.52 & 10.27 & \textcolor{blue}{6.84}\\
            & (0.04) & (0.02) & (0.05) & (0.07) & (0.12) & (0.34) &	(0.06) & (0.04) & (0.06) & (0.12) & (0.04) & (0.69) & (0.04)\\
        \multirow{2}{*}{RDI} & 4.81 & 5.29 & 6.05 & 6.39 & 5.97 & 6.27 & 8.13 & 9.03 & 10.89 & 12.57 & 8.78 & 10.57 & 7.30\\
            & (0.02) & (0.04) & (0.02) & (0.02) & (0.09) & (0.18) &	(0.04) & (0.19) & (0.04) & (0.11) & (0.10) & (0.84) & (0.04) \\
        \multirow{2}{*}{CDR} & 4.79 & 5.16 & 5.75 & 6.23 & 5.64 & 6.05 & 8.20 & 9.23 & 10.81 & 12.25 & 8.97 & 10.60 & 7.20\\
            & (0.04) & (0.03) & (0.03) & (0.10) & (0.03) & (0.26) &	(0.08) & (0.16) & (0.16) & (0.02) & (0.16) & (0.48) & (0.01)\\
        \multirow{2}{*}{D2L} & 4.75 & 5.24 & 5.70 & 6.28 & 5.60 & 6.09 & 8.17 & 9.35 & 10.86 & 12.31 & 9.03 & 10.65 & 7.18\\
            & (0.03) & (0.10) & (0.03) & (0.11) & (0.03) & (0.24) &	(0.05) & (0.11) & (0.09) & (0.01) & (0.21) & (0.62) & (0.12)\\
        \multirow{2}{*}{AUX} & 4.61 & 5.17 & 5.70 & 6.20 & 5.57 & 6.04 & 7.86 & 9.00 & 10.64 & 12.35 & 8.61 & 10.44 & 6.96\\
            & (0.02) & (0.05) & (0.04) & (0.07) & (0.09) & (0.20) &	(0.01) & (0.11) & (0.11) & (0.13) & (0.13) & (0.66) & (0.05) \\
        \multirow{2}{*}{Selfie} & 5.08 & 5.43 & 6.26 & 6.42 & 6.34 & 6.55 & 8.90 & 10.74 & 12.53 & 12.85 & 11.22 & 12.43 & 8.24\\
            & (0.19) & (0.16) & (0.01) & (0.01) & (0.28) & (0.23) &	(1.07) & (0.10) & (0.07) & (0.05) & (0.39) & (0.21) & (0.92)\\
        \multirow{2}{*}{Co-Selfie} & 4.98 & 5.34 & 6.07 & 6.42 & 6.13 & 6.65 & 8.63 & 10.48 & 11.69 & 12.87 & 10.65 & 12.20 & 7.20\\
            & (0.13) & (0.06) & (0.08) & (0.08) & (0.28) & (0.68) &	(0.88) & (0.84) & (0.19) & (0.03) & (0.61) & (0.45) & (0.46)\\
        \multirow{2}{*}{Superloss} & 4.66 & 5.14 & 5.64 & 6.27 & 5.54 & 6.21 & 8.62 & 10.16 & 11.67 & 12.63 & 10.75 & 11.41 & 6.97\\
            & (0.07) & (0.05) & (0.05) & (0.05) & (0.05) & (0.25) &	(0.27) & (0.25) & (0.19) & (0.15) & (1.14) & (0.57) & (0.05)\\
        
        \specialrule{0.7pt}{1pt}{1pt}
        \multirow{2}{*}{\textbf{ConFrag}} & \textcolor{blue}{4.46} & \textcolor{blue}{4.70} & \textcolor{red}{5.04} & \textcolor{blue}{5.84} & \textcolor{red}{5.10} & \textcolor{red}{5.53} & \textcolor{blue}{7.34} & \textcolor{blue}{7.87} & \textcolor{red}{8.89} & \textcolor{red}{11.26} & \textcolor{blue}{8.08} & \textcolor{red}{9.31} & 7.00\\
            & (0.04) & (0.02) & (0.03) & (0.05) & (0.06) & (0.22) &	(0.04) & (0.15) & (0.12) & (0.04) & (0.07) & (0.68) & (0.13)\\
        \multirow{2}{*}{\textbf{Co-ConFrag}} & \textcolor{red}{4.37} & \textcolor{red}{4.66} & \textcolor{blue}{5.07} & \textcolor{red}{5.82} & \textcolor{red}{5.10} & \textcolor{blue}{5.57} & \textcolor{red}{7.09} & \textcolor{red}{7.64} & \textcolor{blue}{8.97} & \textcolor{blue}{11.27} & \textcolor{red}{8.02} & \textcolor{blue}{9.49} & \textcolor{red}{6.58}\\
            & (0.05) & (0.04) & (0.01) & (0.09) & (0.06) & (0.11) &	(0.08) & (0.05) & (0.20) & (0.16) & (0.05) & (0.76) & (0.06)\\
        \bottomrule
    \end{tabular}
    }
    \end{small}
    \end{center}
    \label{tab:main_mae}
\end{table*}

\begin{table*}[t]
    \caption{\textbf{Standard Mean Absolute Error} and its standard deviation to the noise-free trained model on the SHIFT15M-B and MSD-B datasets.
    Lower is better.
    The results are the mean of three random seed experiments.
    Number in parenthesis indicates standard deviation.
    The best and the second best methods are respectively marked in \textcolor{red}{red} and \textcolor{blue}{blue}.
    CNLCU-S/H, Co-Selfie, and Co-ConFrag use dual networks to teach each other as done in \citet{han18coteaching}.
    SPR~\citep{wang22spr} fails to run for SHIFT15M-B due to excessive memory consumption.}
    \begin{center}
    \begin{small}
    \setlength{\tabcolsep}{4.2pt}
    \resizebox{\columnwidth}{!}{%
    \begin{tabular}{lcccccccccccc}
        \toprule
        &\multicolumn{6}{c}{SHIFT15M-B}         &\multicolumn{6}{c}{MSD-B}
        \\\cmidrule(lr){2-7}\cmidrule(lr){8-13}
        &\multicolumn{4}{c}{symmetric}    &\multicolumn{2}{c}{Gaussian} &\multicolumn{4}{c}{symmetric} &\multicolumn{2}{c}{Gaussian}
        \\\cmidrule(lr){2-5}\cmidrule(lr){6-7}\cmidrule(lr){8-11}\cmidrule(lr){12-13} 
        noise rate   & 20 & 40 & 60 & 80 & 30 & 50 & 20 & 40 & 60 & 80 & 30 & 50 \\
        \midrule
        \multirow{2}{*}{Vanilla}            & 7.47 & 8.08 & 8.70 & 9.34 & 7.30 & 7.89 & .5918 & .6475 & .7199 & .7974 & .5848 & .6328 \\
            & (0.02) & (0.07) & (0.08) & (0.04) & (0.05) & (0.08) &	(.0024) & (.0112) & (.0171) & (.0016) & (.0031) & (.0161) \\
        \specialrule{0.1pt}{1pt}{1pt}
        \multirow{2}{*}{CNLCU-S} & 7.74 & 8.18 & 8.51 & 9.21 & 7.90 & 8.28 & .5475 & .5798 & .6644 & .7983 & .5727 & .6151 \\
            & (0.03) & (0.04) & (0.01) & (0.05) & (0.04) & (0.02) &	(.0068) & (.0010) & (.0176) & (.0052) & (.0033) & (.0035) \\
        \multirow{2}{*}{CNLCU-H} & 7.28 & 7.73 & 8.22 & 9.32 & 7.46 & 7.92 & .5483 & .5740 & .6032 & .7098 & .5747 & .5972 \\
            & (0.03) & (0.04) & (0.01) & (0.05) & (0.01) & (0.03) &	(.0034) & (.0027) & (.0055) & (.0065) & (.0040) & (.0019) \\
        \multirow{2}{*}{Sigua} & 7.32 & 7.81 & 8.64 & 9.39 & 7.56 & 8.04 & .5538 & .5861 & .6416 & .8248 & .5839 & .6145 \\
            & (0.03) & (0.03) & (0.06) & (0.24) & (0.02) & (0.08) &	(.0035) & (.0024) & (.0154) & (.0150) & (.0062) & (.0026) \\
        \multirow{2}{*}{SPR} & -&-&-&-&-&-& .5854 & .6462 & .7293 & .7961 & .5741 & .6308 \\
            & (-)&(-)&(-)&(-)&(-)&(-)& (.0059) & (.0048) & (.0147) & (.0113) & (.0028) & (.0119) \\
        \multirow{2}{*}{BMM} & 7.33 & 7.70 & 8.13 & 8.68 & 7.37 & 7.75 & .5649 & .6031 & .6747 & .7849 & .5757 & .6116 \\
            & (0.03) & (0.05) & (0.03) & (0.08) & (0.06) & (0.07) &	(.0044) & (.0081) & (.0099) & (.0073) & (.0016) & (.0025) \\
        \multirow{2}{*}{DY-S} & 7.34 & 7.67 & 8.14 & 8.84 & 7.32 & 7.77 & .5653 & .5908 & .6487 & .7394 & .5728 & .6005 \\
            & (0.02) & (0.03) & (0.04) & (0.10) & (0.03) & (0.10) &	(.0072) & (.0040) & (.0164) & (.0102) & (.0045) & (.0049) \\
        \multirow{2}{*}{C-Mixup} & 7.50 & 7.95 & 8.50 & 9.19 & 7.25 & 7.84 & .5673 & .6185 & .6929 & .7704 & .5630 & .6067 \\
            & (0.03) & (0.03) & (0.03) & (0.05) & (0.04) & (0.08) &	(.0041) & (.0107) & (.0096) & (.0135) & (.0023) & (.0028) \\
        \multirow{2}{*}{RDI} & 7.53 & 8.08 & 8.67 & 9.33 & 7.33 & 7.89 & .6618 & .7113 & .7588 & .8174 & .6517 & .6992 \\
            & (0.02) & (0.02) & (0.05) & (0.03) & (0.02) & (0.08) &	(.0030) & (.0040) & (.0050) & (.0039) & (.0065) & (.0092) \\
        \multirow{2}{*}{CDR} & 7.50 & 8.07 & 8.70 & 9.31 & 7.34 & 7.89 & .5896 & .6444 & .7262 & .7978 & .5836 & .6393 \\
            & (0.05) & (0.03) & (0.05) & (0.02) & (0.04) & (0.08) &	(.0065) & (.0160) & (.0116) & (.0062) & (.0030) & (.0089) \\
        \multirow{2}{*}{D2L} & 7.48 & 8.08 & 8.67 & 9.33 & 7.28 & 7.92 & .5857 & .6559 & .7243 & .8018 & .5769 & .6317 \\
            & (0.03) & (0.07) & (0.07) & (0.01) & (0.01) & (0.13) &	(.0021) & (.0049) & (.0116) & (.0122) & (.0030) & (.0106) \\
        \multirow{2}{*}{AUX} & 7.38 & 8.01 & 8.67 & 9.35 & 7.19 & 7.83 & .5802 & .6462 & .7167 & .7966 & .5753 & .6312 \\
            & (0.01) & (0.04) & (0.04) & (0.04) & (0.04) & (0.08) &	(.0027) & (.0097) & (.0152) & (.0067) & (.0010) & (.0089) \\
        \multirow{2}{*}{Selfie} & 7.18 & 7.55 & 8.37 & 9.46 & 7.23 & 7.64 & .5546 & .5927 & .6574 & .7976 & .6253 & .6787 \\
            & (0.03) & (0.03) & (0.19) & (0.02) & (0.04) & (0.05) &	(.0020) & (.0046) & (.0230) & (.0174) & (.0190) & (.0175) \\
        \multirow{2}{*}{Co-Selfie} & 7.64 & 7.97 & 9.05 & 9.54 & 7.77 & 8.38 & \textcolor{blue}{.5447} & \textcolor{blue}{.5709} & \textcolor{red}{.5923} & .7407 & .5839 & .6187 \\
            & (0.03) & (0.01) & (0.01) & (0.01) & (0.02) & (0.08) &	(.0026) & (.0019) & (.0088) & (.0023) & (.0105) & (.0084) \\
        \multirow{2}{*}{Superloss} & 7.22 & 7.69 & 8.44 & 9.26 & 7.23 & 7.76 & .5460 & .6052 & .6733 & .7959 & .5706 & .6362 \\
            & (0.06) & (0.07) & (0.09) & (0.06) & (0.06) & (0.06) &	(.0036) & (.0104) & (.0153) & (.0405) & (.0065) & (.0140) \\
        \specialrule{0.7pt}{1pt}{1pt}
        \multirow{2}{*}{\textbf{ConFrag}} & \textcolor{blue}{7.02} & \textcolor{blue}{7.27} & \textcolor{red}{7.58} & \textcolor{blue}{8.15} & \textcolor{blue}{7.10} & \textcolor{red}{7.40} & .5499 & .5738 & .6081 & \textcolor{blue}{.6747} & \textcolor{blue}{.5598} & \textcolor{blue}{.5822} \\
            & (0.01) & (0.02) & (0.01) & (0.03) & (0.01) & (0.04) &	(.0035) & (.0039) & (.0051) & (.0098) & (.0050) & (.0084) \\
        \multirow{2}{*}{\textbf{Co-ConFrag}} & \textcolor{red}{6.91} & \textcolor{red}{7.23} & \textcolor{blue}{7.59} & \textcolor{red}{8.14} & \textcolor{red}{7.06} & \textcolor{blue}{7.44} & \textcolor{red}{.5432} & \textcolor{red}{.5631} & \textcolor{blue}{.5941} & \textcolor{red}{.6590} & \textcolor{red}{.5562} & \textcolor{red}{.5796} \\
            & (0.01) & (0.01) & (0.01) & (0.06) & (0.04) & (0.06) &	(.0056) & (.0018) & (.0009) & (.0129) & (.0051) & (.0044) \\
        \bottomrule
    \end{tabular}
    }
    \end{small}
    \end{center}
    \label{tab:main_mae_2}
\end{table*}

\begin{algorithm}[tb]
\caption{Contrastive Fragmentation}\label{alg:fragmented_selection}
    \begin{algorithmic}
        \STATE {\bfseries Input:} Train data $\mathcal{D} = \{\mathcal{X}, Y\}$, Fragment number $F$, KNN parameter $K$, Jitter $J$, Total epochs $N$ 
        \STATE
        \STATE $\Theta = \{\theta_{0,0} \ldots \theta_{i,j}\}$ \COMMENT{feature extractors}
        \STATE $\Phi = RandomInit()$ \COMMENT{regression model}
        \STATE
        \STATE $\mathcal{D}_{1 \ldots F} = Fragmentation(\mathcal{D})$ \COMMENT{\S~\ref{subsec:fragmentation}. 1}
        \STATE $\mathcal{P} = ContrastivePairing(\mathcal{D}_{1 \ldots F})$ \COMMENT{\S~\ref{subsec:fragmentation}. 2$\sim$4}
        \FOR {$n$ {\bfseries to} $N$}
            \STATE $\text{\color{blue}{\# train feature extractors}}$
            \STATE $\mathcal{P}^{jitter} = NeighborhoodJittering(\mathcal{D}, F, J)$ \COMMENT{neighborhood jittering (\S~\ref{sec:jittering})}
            \FOR {$(\mathcal{D}_i^{jitter}, \mathcal{D}_j^{jitter})$ {\bfseries in} $\mathcal{P}^{jitter}$}
                \STATE $\mathcal{D}_{i,j}^{jitter} = \mathcal{D}_i^{jitter} \cup \mathcal{D}_j^{jitter}$
                \STATE $\text{train } p(f; \theta_{i,j},\mathcal{D}_{i,j}^{jitter})$ 
            \ENDFOR
            \STATE
            \STATE $\color{blue}{\text{\# initialize } \mathcal{S}, \mathcal{S}^p, \mathcal{S}^r}$
            \STATE $\mathcal{S}, \mathcal{S}^p, \mathcal{S}^r = \{\}, \{\}, \{\}$ \COMMENT{selected samples}
            \STATE
            \STATE $\color{blue}{\text{\# obtain } \mathcal{S}^p, \mathcal{S}^r}$
            \FOR{$(x, y)$ {\bfseries in} $\mathcal{D}$}
                \FOR{$f=1$ {\bfseries to} $F$}
                    \STATE $\text{calculate }\rho^\text{}_f(y)$ \COMMENT{fragment prior (Eq.~\ref{eq:np})}
                    \STATE $\text{\color{blue}{\# use two types of classification for neighborhood agreement}}$
                    \STATE $\text{calculate }\alpha_f^p(x; \mathcal{D}_{1 \ldots F}, \Theta)$ \COMMENT{predictive neighborhood agreement (Eq.~\ref{eq:na_final})}
                    \STATE $\text{calculate }\alpha_f^r(x; \mathcal{D}_{1 \ldots F}, \Theta)$ \COMMENT{representational neighborhood agreement (Eq.~\ref{eq:na_final})}
                \ENDFOR
                \STATE $p^p(s|x,y, \mathcal{D}_{1 \ldots F};\Theta) = \sum_{f}^{F} \rho^\text{}_f(y)\alpha_f^p(x; \mathcal{D}_{1 \ldots F}, \Theta)$ \COMMENT{pred. sample probability (Eq.~\ref{eq:mcf})}
                \STATE $p^r(s|x,y, \mathcal{D}_{1 \ldots F};\Theta) = \sum_{f}^{F} \rho^\text{}_f(y)\alpha_f^r(x; \mathcal{D}_{1 \ldots F}, \Theta)$ \COMMENT{repr. sample probability (Eq.~\ref{eq:mcf})}

                \STATE $\text{sample } \{u^p, u^r\} \sim uniform(0,1)$
                \IF {$p^p(s|x,y, \mathcal{D}_{1 \ldots F};\Theta) > u^p$}
                \STATE $\mathcal{S}^p = \mathcal{S}^p \cup (x, y)$
                \ENDIF
                \IF {$p^r(s|x,y, \mathcal{D}_{1 \ldots F};\Theta) > u^r$}
                \STATE $\mathcal{S}^r = \mathcal{S}^r \cup (x, y)$
                \ENDIF
            \ENDFOR
            \STATE
            \STATE $\color{blue}{\text{\# union filtered samples } (\mathcal{S}^p, \mathcal{S}^r)}$
            \STATE $\mathcal{S} = \mathcal{S}^p \cup \mathcal{S}^r$
            \STATE
            \STATE $\text{\color{blue}{\# train regression model}}$
            \STATE $\Phi = TrainOneEpoch(S ; \Phi)$
            \STATE 
        \ENDFOR
    \end{algorithmic}
\end{algorithm}

\clearpage

\end{document}